\documentclass[10pt,letterpaper]{article}
\usepackage[top=0.85in,left=2.75in,footskip=0.75in]{geometry}

\usepackage{amsmath,amssymb}

\usepackage{changepage}

\usepackage{textcomp,marvosym}

\usepackage{cite}

\usepackage[ruled,noline, linesnumbered]{algorithm2e} 

\usepackage{nameref,hyperref}

\usepackage[right]{lineno}

\usepackage{float}

\usepackage[nopatch=eqnum]{microtype}
\DisableLigatures[f]{encoding = *, family = * }

\usepackage[table]{xcolor}

\usepackage{array}
\usepackage{booktabs}

\usepackage{orcidlink}

\newcolumntype{+}{!{\vrule width 2pt}}

\newlength\savedwidth

\raggedright
\usepackage[aboveskip=1pt,labelfont=bf,labelsep=period,justification=raggedright,singlelinecheck=off]{caption}

\makeatletter
\renewcommand{\@biblabel}[1]{\quad#1.}
\makeatother

\usepackage{lastpage,fancyhdr,graphicx}
\usepackage{epstopdf}
\usepackage{mathtools}
\fancyheadoffset[L]{2.25in}
\fancyfootoffset[L]{2.25in}
\usepackage{todonotes}
\newboolean{SHOWCOMMENTS}
\setboolean{SHOWCOMMENTS}{true}
\ifthenelse{\boolean{SHOWCOMMENTS}}{
\newcommand{\tdHZ}[1]
{\todo[color=green!25,inline]{\footnotesize{\bf Henrik:} #1}}
}{
\newcommand{\tdHZ}[1]{}
}

\ifthenelse{\boolean{SHOWCOMMENTS}}{
\newcommand{\tdMK}[1]
{\todo[color=red!25,inline]{\footnotesize{\bf Martin:} #1}}
}{
\newcommand{\tdMK}[1]{}
}

\usepackage[capitalise]{cleveref}
\Crefname{figure}{Fig}{Figs}

\newcommand{\mbf}[1]{{\mathbf{#1}}}

\newcommand{\Tcl}{T_{\mathsf{cl}}}

\ifthenelse{\boolean{SHOWCOMMENTS}}{
\newcommand{\tdRK}[1]
{\todo[color=blue!25,inline]{\footnotesize{\bf Raouf:} #1}}
}{
\newcommand{\tdRK}[1]{}
}

\newtheorem{theorem}{Theorem}

\newtheorem{definition}{Definition}

\def\1{\bm{1}}

\def\gD{{\mathcal{D}}}
\def\gE{{\mathcal{E}}}
\def\gM{{\mathcal{M}}}
\def\gN{{\mathcal{N}}}
\def\gR{{\mathcal{R}}}

\def\sR{{\mathbb{R}}}

\newcommand{\out}[1]{}

\begin{document}
\vspace*{0.2in}

\begin{flushleft}
{\Large
\textbf\newline{Differentially private federated learning for localized control of infectious disease dynamics} 
}
\newline
\\
Raouf Kerkouche\orcidlink{0000-0002-1458-7805}\textsuperscript{1}\Yinyang,
Henrik Zunker\orcidlink{0000-0002-9825-365X}\textsuperscript{2}\Yinyang,
Mario Fritz\orcidlink{0000-0001-8949-9896}\textsuperscript{1},
Martin J. Kühn\orcidlink{0000-0002-0906-6984}\textsuperscript{2,3}
\\
\bigskip
\textbf{1} CISPA Helmholtz Center for Information Security, Saarbrücken, Germany
\textbf{2} Institute of Software Technology, Department of High-Performance Computing, German Aerospace
Center, Cologne, Germany, 
\textbf{3} Bonn Center for Mathematical Life Sciences and Life and Medical Sciences Institute, University of Bonn, Bonn, Germany
\bigskip

%
%
\Yinyang\; Shared first author. These authors contributed equally to this work.




* Martin.Kuehn@DLR.de

\end{flushleft}
\section*{Abstract}
In times of epidemics, swift reaction is necessary to mitigate epidemic spreading. For this reaction, localized approaches have several advantages, limiting necessary resources and reducing the impact of interventions on a larger scale. However, training a separate machine learning (ML) model on a local scale is often not feasible due to limited available data. Centralizing the data is also challenging because of its high sensitivity and privacy constraints. In this study, we consider a localized strategy based on the German counties and communities managed by the related local health authorities (LHA). For the preservation of privacy to not oppose the availability of detailed situational data, we propose a privacy-preserving forecasting method that can assist public health experts and decision makers. ML methods with federated learning (FL) train a shared model without centralizing raw data.  Considering the counties, communities or LHAs as clients and finding a balance between utility and privacy, we study a FL framework with client-level differential privacy (DP). We train a shared multilayer perceptron on sliding windows of recent case counts to forecast the number of cases in the future, while clients exchange only norm-clipped updates and the server aggregated updates with DP noise. We evaluate the approach on COVID-19 data on county-level during two phases-November 2020 and March 2022 (Omicron). As expected, very strict privacy ($\varepsilon\le0.5$) yields unstable, unusable forecasts. At a moderately strong but still privacy-preserving level ($\varepsilon = 2$), the DP model closely approaches the non-DP model: $R^2\approx 0.94$ (vs. 0.95) and mean absolute percentage error (MAPE) $\approx 26\%$ in November 2020; $R^2\approx 0.88$ (vs. $0.93$) and MAPE $\approx 21\%$ in March 2022. Overall, client-level DP-FL can deliver useful county-level predictions with strong privacy guarantees, and viable privacy budgets depend on epidemic phase, allowing privacy-compliant collaboration among health authorities for local forecasting.

\section*{Author summary}
We address a practical challenge that many local health authorities face during fast accelerating outbreaks: they hold timely, sensitive case data that could improve short‑term forecasts if they are used. However, legal and ethical constraints often prevent central sharing. We studied if several local health authorities (LHAs) can jointly train one central forecasting model while each LHA keeps its data entirely local. Hereby, we build on federated learning (FL), where every site trains a shared neural network on its own recent case counts and only sends model updates. To further protect contributors, we apply differential privacy (DP), adding random noise so that no single participant's information can be inferred from the final model. We test our approach on COVID‑19 data published on county-level and for two different phases of the pandemic. We observe the expected tradeoff: very strict privacy settings destroy forecast quality, whereas moderately strong privacy-preservation approaches still yield trajectories close to a non-DP baseline. Our findings show that a privacy‑preserving approach for regional infectious disease forecasting is feasible, and how an appropriate privacy level should be derived for different epidemic phases to maintain both usefulness and protection.


\section*{Introduction}

In epidemic emergencies or situations of seasonally peaking endemic pathogens, public health experts and decision makers can benefit from predictions of infectious disease dynamics to proactively react to upcoming challenges. For prediction tasks, mechanistic mathematical models as well as ML models have proven to provide reliable outcomes. While for mechanistic population~\cite{mohring_estimating_2024,merkt_long-term_2024,kheifetz_modeling_2025}, metapopulation~\cite{zunker_novel_2024, ZUNKER2025116782}, agent-based~\cite{kerr_covasim_2021,muller_explicit_2023,KERKMANN2025110269} or combined differential equation- and ML-based~\cite{radev_outbreakflow_2021,schmidt2025} approaches, new situations might require the adaptations of existing mechanisms, many ML models are developed in a purely data-driven fashion, reducing the number of manual interventions cycles. In particular, classical simulation or forecasting studies~\cite{muller_explicit_2023,mohring_estimating_2024,merkt_long-term_2024,kheifetz_modeling_2025,KERKMANN2025110269} mostly used publicly available datasets and did not consider the aspect of privacy-preservation.

The traditional approach to training machine learning models typically requires centralizing data from multiple parties. This creates serious privacy risks when handling sensitive information such as medical records, which often cannot be shared directly. In infectious disease contexts, these records include precise dates of symptom onset (essential for estimating reporting delays), assumed transmission events, and detailed demographics maintained by hospitals or LHAs. These features are critical for predicting future cases but are removed from public datasets to preserve anonymity. While aggregation reduces disclosure risk, it removes these valuable signals and does not eliminate the risk of re-identification~\cite{dwork2017exposed}. Ensuring privacy-preserving approaches is therefore essential when working with such data.

To train ML models on spatially distributed data, FL has been introduced, enabling the collaborative training of a shared model without the need for data centralization~\cite{pmlr-v54-mcmahan17a, Rieke2020, Dayan2021}. FL allows each client to train the model on their own data and share only the model updates, or gradients, through a central server. The server aggregates these updates to improve the shared model, which is then redistributed to participants for further refinement. This process repeats until the model achieves satisfactory performance.

FL is grounded in three key objectives: First, it aims to protect the privacy of each participant's data by exchanging model updates rather than the data itself. Second, it seeks to lower communication costs by allowing several local training iterations before the exchange of updates. Finally, by involving only a subset of clients in each round, it reduces communication overhead and increases resilience to the temporary unavailability of clients, offering particular advantages in scenarios with a large number of clients.

However, sharing gradients during collaborative learning can still reveal sensitive information from the training data of individual parties. Advanced attacks have demonstrated that adversaries can infer specific data records or group characteristics from intercepted model updates~\cite{VFLllgusenix, li2021label, Property, nasr2019comprehensive,LLG, IDLG, DLG, Shiri2024}. In extreme cases, they can reconstruct entire training samples from the gradients~\cite{IDLG, DLG}. Therefore, merely avoiding the sharing of actual training data can not completely avoid privacy leakage.

Initial works on calibrating noise to sensitivity~\cite{dwork2006calibrating} led to an extensive work of algorithmic foundations of \textit{Differential Privacy}~\cite{dwork2014algorithmic}. DP has emerged as a preeminent framework for enhancing data privacy, countering various privacy attacks associated with FL. DP provides strong privacy assurances by ensuring that the behavior of the aggregated model does not depend significantly on any single client's data. Instead, it captures commonalities across all contributions. In the context of federated learning, several foundational works have combined $(\varepsilon,\delta)$-DP with standard federated learning algorithms that aggregate client updates on a central server~\cite{geyer2018differentiallyprivatefederatedlearning,9069945,Abadi_2016}, or with local differential privacy mechanisms in which each client perturbs its model update before sending it to the server~\cite{10.1145/3378679.3394533}. An overview of the federated learning landscape, including privacy preservation, is given by~\cite{kairouz2021advancesopenproblemsfederated}. Implementing DP in these settings typically involves clipping client updates to a fixed sensitivity and adding Gaussian noise during aggregation, which helps protect participant-specific information while preserving overall learning utility. In 2021, the authors of~\cite{ficek21} had observed a substantial increase in the use of DP in health research publications and, among others, the authors of~\cite{kerkouche2021privacy, rahimian2022practical, ryu_appfl_2022,madduri24} developed a framework for privacy-preserving FL. In the context of the COVID-19 pandemic, the authors of~\cite{hauer21} raised the awareness for a potential distortion of COVID-19 death rates when using DP with too small dataset groupings. During the COVID-19 pandemic, more than 1.9 million COVID-19 cases across Europe have been integrated in the DataSHIELD (Data Aggregation Through Anonymous Summary-statistics from Harmonised Individual-levEL Databases) platform~\cite{wilson_datashield_2017}, to allow for FL approaches to derive potential strategies against COVID-19~\cite{penalvo_unravelling_2021}. Among other things, DataSHIELD has already been used for other infectious diseases, e.g., to study malaria intervention in Mozambique~\cite{huth_federated_2024}.

In order to report dynamics and allow predictions by simultaneously addressing privacy concerns, COVID-19 case counts had been aggregated and reported on county-level throughout the pandemic in Germany~\cite{Robert_Koch-Institut_SARS-CoV-2_Infektionen_in_2025}. We use these official aggregates as a proof-of-concept dataset to evaluate our method, which is intended for settings where finer-grained, potentially sensitive data remain with the LHAs. While aggregation reduces disclosure risk, it does not provide formal privacy guarantees~\cite{dwork2017exposed}. 
Our work builds on established algorithms for federated learning with differential privacy~\cite{geyer2018differentiallyprivatefederatedlearning,10.1145/3378679.3394533,9069945} and selected DP-FL mechanism. To allow the reliable application in a public health context, we focus on a realistic infectious disease forecasting use case to study and evaluate client-level DP-FL on county-level COVID-19 data from two distinct epidemic phases in Germany. Selecting an established setup in which clients send clipped model updates and the server adds Gaussian noise calibrated to a target $(\varepsilon,\delta)$-DP guarantee via a R\'enyi DP accountant, we map out the privacy-utility tradeoff across a broad grid of privacy budgets and identify ranges of $\varepsilon$ for which forecasts remain practically useful. Our findings are intended to inform public health researchers and practitioners about which privacy budgets are compatible with accurate localized predictions.

\section*{Materials and methods}
We develop a FL scheme with client-level DP to produce short-term local case forecasts while keeping raw data decentralized.
Our goal is to develop a solution that ensures client-level privacy. For instance, in scenarios involving collaborative LHAs, our goal is to safeguard any information unique to each LHA's training data. An adversary should not be capable of discerning from the received model or its updates whether any client’s data has been included in the federated run (within $\varepsilon$ and $\delta$ bounds). DP provides plausible deniability not only for groups of a client's samples but also for any client participating in the federated run. Consequently, any adverse privacy impact on clients (or their training samples) cannot be directly linked to their participation in the protocol run.

\subsection*{Data preprocessing}
Our objective is time‐series forecasting of local case counts using data which is reported on local level. We construct supervised learning examples by using a historical window of size $H=10$ days over each administrative unit’s time series and predicting the case count $P=7$ days ahead. The window size $H$ and prediction horizon $P$ were determined empirically to ensure robust predictive performance.
Let $c_{k,t}$ denote the raw, unnormalized number of daily reported cases for geographic unit $k$, $k=1,2,3,\ldots$ on day $t$. Therefore, the input vector
\begin{align}\label{eq:input}
\mathbf{x}_{k,t} = \bigl[c_{k,t-H+1},\dots,\,c_{k,t}\bigr]
\end{align}
contains the reported daily cases on the preceding $H$ days, and the target
\begin{align}\label{eq:output}
y_{k,t} = c_{k,t+P}
\end{align}
is the case count $P$ days into the future. Missing reports are handled as zero reported cases. To mitigate weekly reporting artifacts (e.g., inconsistent weekend reporting), we applied a centered 7-day moving average. This smoothing is crucial not only for noise reduction but also to stabilize gradient updates in the FL process.
For model evaluation, we construct examples by using a historical window of $H = 10$ consecutive days and predicting the case count $P = 7$ days ahead for each county. Within each epidemic period (November~2020 and March~2022) and for each county separately, we order all resulting input-label pairs by the date of the prediction target and assign the earliest $90~\%$ to the training set and the most recent $10~\%$ to the test set. In this way, all test samples correspond to later days than the training samples for the same county, reflecting the intended short-term forecasting use case. The November~2020 and March~2022 datasets are handled in separate experiments, i.e. models are trained and evaluated independently on each period. After splitting, we remove the location identifier from the feature vectors since they should not influence the model in any way.

\subsection*{Model architecture}
We use PyTorch~\cite{Ansel_PyTorch_2_Faster_2024} to build a fully connected multilayer perceptron (MLP) as our prediction model.
The architecture is consistent across both spatial resolutions, community- and county-based; see the \textit{Results} section for details on the datasets used.
The input to the network is a vector of length $H$ containing the $H$ most recent daily case counts; with zero reported cases counting as daily case count. This vector is propagated through three fully connected hidden layers of sizes 128, 64, and 32 neurons, each followed by a rectified linear unit (ReLU) activation function. A final linear output unit produces the predicted case count $P$ days ahead. We use the mean squared error (MSE) as the loss function, which is minimized by the Adam optimizer with a learning rate of 0.001.

\subsection*{Differential privacy with federated learning}
We use a federated learning approach with client-level differential privacy. To formally define our privacy guarantees, we rely on the standard definition of $(\varepsilon, \delta)$-differential privacy~\cite{dwork2014algorithmic}. In the following, we briefly summarize the background of differential privacy with our model choices.

    \subsection*{Differential Privacy}
    \begin{definition}[Differential Privacy~\cite{dwork2014algorithmic}] A randomized mechanism $\gM$ with
    range $\gR$ satisfies $(\varepsilon, \delta)$-differential privacy, if for any two adjacent datasets $D$ and $ D'$, i.e., $D'=D \cup \{x\}$ for some $x$ in the data domain (or vice versa), and for any subset of outputs $O \subseteq \gR$, it holds that 
    \begin{equation}
        \Pr[\gM(D) \in O] \leq e^\varepsilon \Pr[\gM(D') \in O] + \delta
    \end{equation}
    \end{definition}
    
    Intuitively, DP guarantees that an adversary, provided with the
    output of $\mathcal{M}$, can draw almost the same conclusions (up to $\varepsilon$ with probability larger than $1 - \delta$) about any group no matter if it is included in the input of $\mathcal{M}$ or not~\cite{dwork2014algorithmic}. This means, for any group owner, a privacy breach is unlikely to be due to its participation in the dataset.
    
    We use the Gaussian mechanism to upper bound privacy leakage when transmitting information from clients to the server.
    
    \begin{definition} (Gaussian Mechanism~\cite{dwork2014algorithmic})
    Let $f: \sR^n \rightarrow \mathbb{R}^d$ be an arbitrary function that maps $n$-dimensional input to $d$ logits with sensitivity being:
     \label{def:sensitivity}
    \begin{equation}
      S= \max_{D,D'} \Vert f(D)-f(D')\Vert_2  
    \end{equation}
    over all adjacent datasets $D$ and $D' \in \gD$.
    The Gaussian Mechanism  $\mathcal{M}_\sigma$, parameterized by $\sigma$, adds noise into the output, i.e.,
    \begin{equation}
      \mathcal{M_\sigma}(x) = f(x) + \mathcal{N}(0,\sigma^2 I).
    \end{equation}
    \label{def:gaussian_mechanism}
    \end{definition}
    
    \noindent \textbf{Adversarial Model.} We assume an adversary that aims at inferring private information of the participating clients' training datasets. We consider a strong adversary who can control a certain fraction of the participating clients. The adversary can access the training data, including all the parameters of the local models of the corrupted clients (i.e., it has white-box access to their local models). However, we consider that our server is trusted. 
    The adversary is \emph{passive} (i.e., honest-but-curious), that is, it follows the learning protocol faithfully. \\
    
    \noindent \textbf{Privacy Model.} In federated learning, the notion of \textit{adjacent (neighboring) datasets} used in DP generally refers to pairs of datasets differing by one client (\textit{client-level} DP), or by one group of one user (\textit{group-level} DP), or by one data point of one user (\textit{record-level} DP). Our approach focuses on the \textit{client-level} DP, where each client corresponds to a data holder, such as a Local Health Authority (LHA). Therefore, we aim to develop a solution that ensures client-level privacy. For instance, in scenarios involving collaborative LHAs, our goal is to safeguard any information unique to each LHA's training data. The adversary should not be capable of discerning from the received model or its updates whether any client’s data has been included in the federated run (within $\varepsilon$ and $\delta$ bounds). We believe this adversarial model applies in many practical scenarios where private information extends across multiple samples within a client's training data (for example, the presence of a group of samples, such as individuals with a specific feature). Differential Privacy provides plausible deniability not only for groups of a client's samples but also for any client participating in the federated run. Consequently, any adverse privacy impact on clients (or their training samples) cannot be directly linked to their participation in the protocol run. The detailed privacy analysis, including the accounting via R\'enyi Differential Privacy (RDP), is provided in~\nameref{S1_text}.

\subsubsection*{DP-FL Algorithm}
The core methodology employs FL with client-level DP, summarized in~\cref{alg:fl_loki_dp}. Our client-level DP-FL protocol is conceptually aligned with established DP-FL schemes~\cite{geyer2018differentiallyprivatefederatedlearning,9069945}, where each selected client clips its local update to a fixed sensitivity, and the server adds Gaussian noise to the averaged update with a noise multiplier chosen via the R\'enyi DP accountant.
Following standard practices in differentially private federated learning~\cite{kerkouche2021privacy}, the clipping threshold $S$ and the client sampling rate $q$ were determined empirically.
All hyperparameters are additionally explained in~\cref{tab:hyperparams}. The training process basically follows four steps:
\begin{enumerate}
    \item \textbf{Client selection:} In each round $t$, a subset $\mathbb{K}$ of clients is selected from the total number of clients $N_{\mathrm{total}}$. Specifically, each client is independently included with probability 
    \[
    q \;=\; \frac{m}{N_{\mathrm{total}}},
    \]
    so that on average $m$ participants (i.e., geographic units) contribute updates per round. This random sampling strategy reduces both communication and computational load. It also represents slow or unavailable clients, since each round depends only on the selected clients.  When DP is enabled, this subsampling also amplifies privacy by lowering the effective per-round privacy loss.  
    
    \item \textbf{Local training:} Then, each selected client $k \in \mathbb{K}$ receives the current global model weights $\mathbf{w}_{i-1}$. The client then performs local training for a fixed number of epochs $E$ on its own local dataset which remains local at all times. After local training completes, resulting in local model weights $\mathbf{w}_{i-1}^k$, the client $k$ computes the local model update $\Delta \mathbf{w}_i^k = \mathbf{w}_{i-1}^k - \mathbf{w}_{i-1}$ relative to the initial given global model. Only this update is sent to the server, ensuring that no privacy critical data leaves the local client.
    
    \item \textbf{DP mechanism:} When client-level DP is enabled (i.e.,\ $\varepsilon<\infty$), each participating client first applies the Euclidean 2-norm bound on its model update $\Delta \mathbf{w}_i^k$ by clipping to a fixed sensitivity $S=0.5$:  
    \[
    \Delta \widehat{\mathbf{w}}_i^k \;=\;\frac{\Delta \mathbf{w}_i^k}{\max\!\bigl(1,\|\Delta \mathbf{w}_i^k\|_2/S\bigr)}.
    \]  
    The server then aggregates these clipped updates by summation and divides by the expected batch size $m$ to obtain the mean update. Finally, Gaussian noise $\mathcal{N}(0,\sigma^2 \mathbf{I})$ is added to this averaged update, where  
    \[
    \sigma \;=\;\frac{S\cdot c}{m},
    \]  
    and the noise multiplier $c$ is chosen via the Rényi DP (RDP) accountant~\cite{mironov2017renyi}. If privacy is disabled ($\varepsilon=\infty$), no clipping or noise is applied and the raw updates $\Delta \mathbf{w}_i^k$ are simply averaged.
    
    \item \textbf{Global model update:} 
    Let
    \[
    \Delta\overline{\mathbf{w}}_i \;:=\; \frac{1}{m}\sum_{k\in\mathbb{K}} \Delta \widehat{\mathbf{w}}_i^{\,k}
    \]
    denote the mean clipped update.
    In the DP case, we draw $\xi_i \sim \mathcal{N}(0,\sigma^2 \mathbf{I})$. Then, the server updates the global model by adding the final averaged update:
    \[
    \mathbf{w}_i = \mathbf{w}_{i-1} + \Delta\overline{\mathbf{w}}_i + \xi_i
    \]
    In the non-DP case, this reduces to:
   $\mathbf{w}_i = \mathbf{w}_{i-1} + \Delta\overline{\mathbf{w}}_i$.

\end{enumerate}
After $\Tcl$ rounds, the final global model is evaluated on the test dataset.
To account for variability due to real random client selection and noise addition, the entire training steps are repeated multiple times with different random seeds.

\begin{algorithm}[H]
\small
		\caption{Federated learning with client-level privacy.}
        \label{alg:fl_loki_dp}
	\DontPrintSemicolon
	{\bf Server:}\;
	\Indp Initialize common model $\mbf{w}_0$\;
	\For {$i=1$ \KwTo $\Tcl$}
	{
	    Select a subset $\mathbb{K}$ of clients randomly with probability $q$\;
        Initialize list of updates $U = \mathbf{[}\,\mathbf{]}$\;
		\For {each client $k\in\mathbb{K}$}
		{	
			$\Delta \widehat{\mbf{w}}_i^k  = \mathbf{Client}_k(\mbf{w}_{i-1})$\;
            Insert $\Delta \widehat{\mbf{w}}_i^k$ into $U$\;
		}
		$\Delta\overline{\mbf{w}}_i = \frac{1}{m} \sum_{\Delta \widehat{\mbf{w}} \in U} \Delta \widehat{\mbf{w}}$\; 
        \uIf{DP enabled}{
            Calculate noise multiplier $c$ using an RDP accountant for $(\varepsilon, \delta, q, \Tcl, E)$\;
            $\sigma = (S \cdot c) / m$\;
            Draw $\xi_i \sim \mathcal{N}(0, \sigma^2 \mathbf{I})$\;
            $\mbf{w}_{i} = \mbf{w}_{i-1} + \Delta\overline{\mbf{w}}_i + \xi_i$\;
        }
        \Else
        {
             $\mbf{w}_{i} = \mbf{w}_{i-1} + \Delta\overline{\mbf{w}}_i$\;
        }
	}
    \Indm {\bf $\mathbf{Client}_{k}(\mbf{w_{i-1}})$:}\;
    \Indp
	$\widetilde{\mbf{w}}_{i} = \mbf{w}_{i-1}$\; 
    Train $\widetilde{\mbf{w}}_{i}$ on local data $D_k$ for $E$ epochs using Adam ($\eta=0.001$)\;
    $\Delta \mbf{w}_i^k = \widetilde{\mbf{w}}_{i} - \mbf{w}_{i-1}$\;
	$\Delta \widehat{\mbf{w}}_i^k = \Delta \mbf{w}_i^k / \max\left(1, \frac{||\Delta \mbf{w}_i^k||_2}{S}\right)$\;
    \textbf{Output:} $\Delta \widehat{\mbf{w}}_i^k$\;
\end{algorithm}

\begin{table}[H]
  \centering
  \caption{\textbf{Overview of model hyperparameters}}
  \label{tab:hyperparams}
  \begin{tabular}{@{}l p{0.7\textwidth}@{}}
    \toprule
    \textbf{Symbol} & \textbf{Description} \\
    \midrule
    $H$              & Input window length (days)\\
    $P$              & Prediction horizon (days)\\
    $E$              & Number of local training epochs per round \\
    $m$              & Predefined expected number of clients selected per round (expected batch size) \\
    $\Tcl$ & Total number of FL rounds \\
    $q$              & Probability of selecting each client per round\\
    $S$              & Euclidean 2-norm clipping bound for each update \\
    $c$              & Noise multiplier for DP Gaussian mechanism \\
    $\varepsilon$    & DP budget \\
    $\delta$         & Privacy failure probability \\
    $\eta$           & Learning rate (Adam) \\
    \bottomrule
  \end{tabular}
\end{table}

\section*{Results}

In this section, we provide results of our DP-FL approach applied to county-level data and discuss potential extensions and preliminary results for finer resolved data.

\subsection*{Dataset}
\begin{figure}[h!]
    \centering\includegraphics[width=0.9\linewidth]{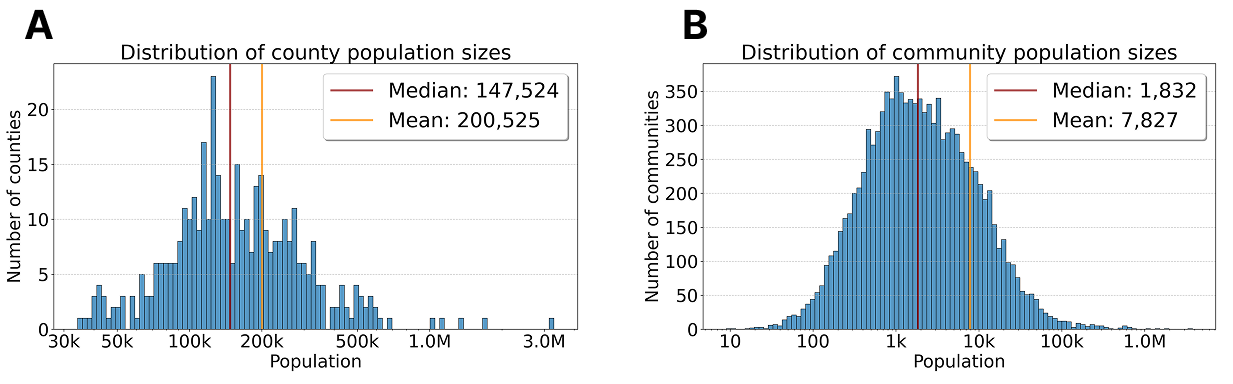}
    \caption{\textbf{Distribution of population sizes on community and county-level.} Panel A shows the distribution of county population sizes across all 400 German counties, with a median population of 147,524 and mean population of 200,525. Panel B displays the distribution of community population sizes across the 10,786 communities included in our analysis, with a substantially smaller median population of 1,832 and mean of 7,827. Red and orange vertical lines indicate median and mean values, respectively.}
    \label{fig:community_vs_county}
\end{figure}

We use official reported county-level COVID-19 case data from the LHAs transferred to the Robert Koch Institute (RKI)~\cite{Robert_Koch-Institut_SARS-CoV-2_Infektionen_in_2025} and accessed via the \textit{memilio-epidata} Python package~\cite{Bicker_MEmilio_v2_0_0_-}. We consider two different pandemic stages, first, late-stage as of March 2022 and, second, early stage as of November 2020. Throughout this work, the November~2020 data and March~2022 data refer to two separate one-month windows of the county-level COVID-19 cases data. Models are trained and evaluated independently on each period.
For each county and day, we obtain the newly confirmed infections plus estimated recoveries and deaths reported with the (assumed) day of infection. This data covers all 400 German counties. While the published data is stratified on county-level, the counties can, again, be divided by communities, summing up to a total of 10,786 communities in Germany. In~\cref{fig:community_vs_county}, Panel A, shows the population distribution across the counties, highlighting the variation in population sizes. The median county population is 147,524, while the mean population is 200,525, as several counties have a substantial larger population. The largest county by population has 3,685,265 inhabitants (Berlin, after merging all Berlin districts), and the total population across all counties is approximately 83.6 million. This heterogeneity in county sizes presents a particular challenge for our FL approach. Counties with larger populations typically report higher absolute case numbers. In addition, we show in Panel B the distribution of population sizes on community-level.

\subsubsection*{Synthetic community data}
To examine whether predictive performance is maintained on finer levels and to evaluate robustness, we generated synthetic case datasets on community-level in Germany. In~\cref{fig:community_vs_county}, we show the distribution of population sizes for community and county sizes. While the median county population is 147,524 as of 2022, the median community population is 1,832; see~\cite{population_statistic}. 

\begin{figure}[h!]
    \centering\includegraphics[width=0.7\linewidth]{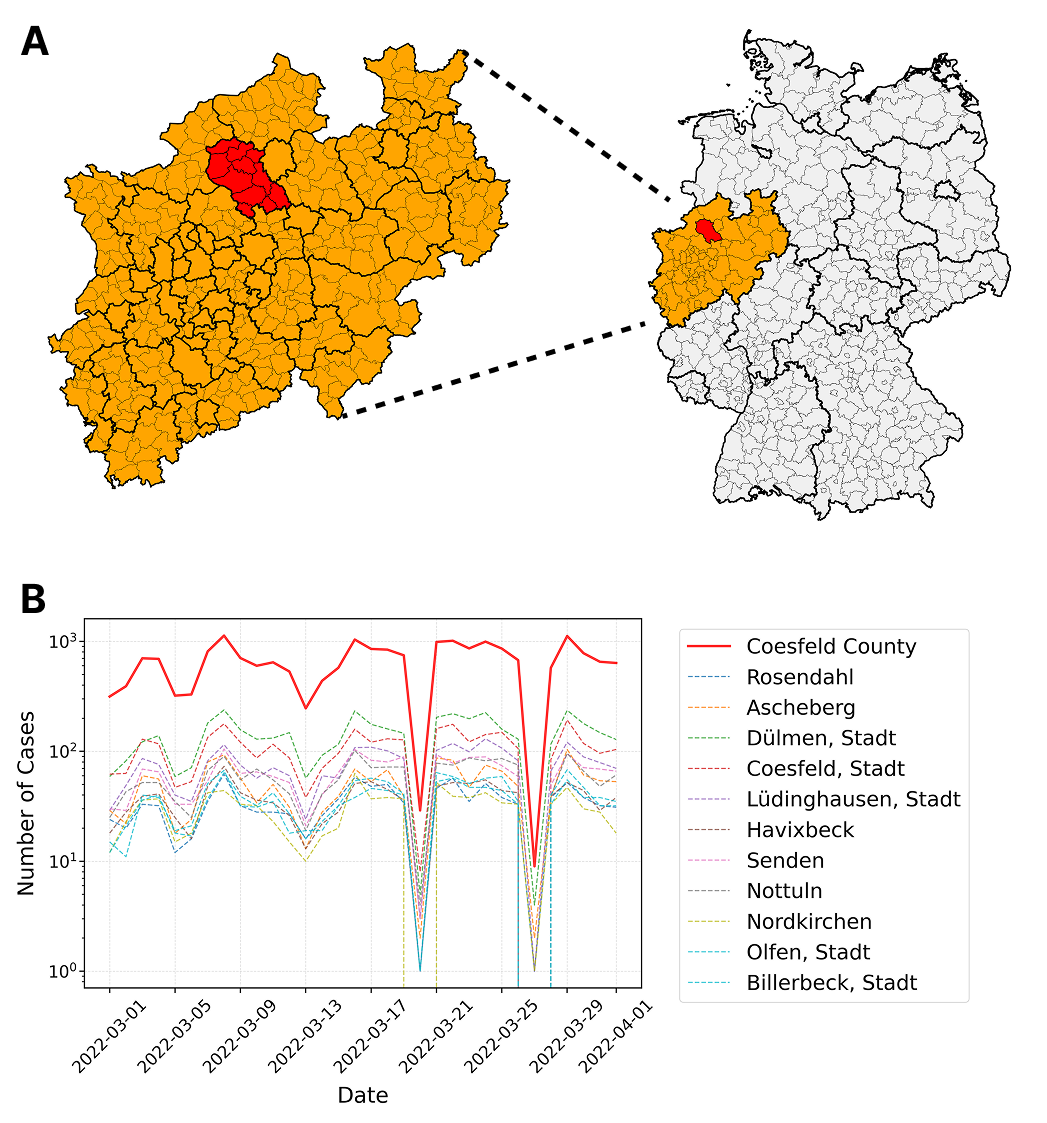}
    \caption{\textbf{Spatial resolution of German counties and communities and community-based case data.} Panel A presents two maps shown side by side. The left map presents North Rhine-Westphalia (NRW) stratified into counties (thick black boundaries) and communities (thin gray boundaries), highlighting county "Coesfeld" in red. The right map displays the stratification of Germany into federal states (thick black boundaries) and counties (thin gray boundaries), highlighting the federal state NRW in orange. Panel B shows the obtained community-based dataset for "Coesfeld" for March 2022, with the county-level aggregate shown in red and individual community trajectories in various colors. The left map from Panel A is using geodata ``Verwaltungsgebiete 1:250 000 (VG250)'' from BKG (2026) dl-de/by-2-0, Data sources: \url{https://sgx.geodatenzentrum.de/web_public/gdz/datenquellen/datenquellen_vg_nuts.pdf}.}
    \label{fig:coesfeld}
\end{figure}

To generate a community-based dataset, we sampled randomly from the county-based dataset, using weights given by community population divided by the corresponding county population. For the spatial stratification and an exemplary result for the county of "Coesfeld", see~\cref{fig:coesfeld}, Panel B. In the resulting dataset, zeros were imputed for days when no case data was sampled for the particular community.

Based on this synthetic dataset, we applied our DP-FL approach. Instead of reporting the results directly, which measured in MAPE or $R^2$ looked much better than they actually were, we provide preliminary analyses on the outcomes. In particular, when DP was used (noise added to the aggregated, clipped updates), we did not observe deteriorated predictions, a precautionary signal.

\begin{figure}[h!]
    \centering
    \includegraphics[width=\linewidth]{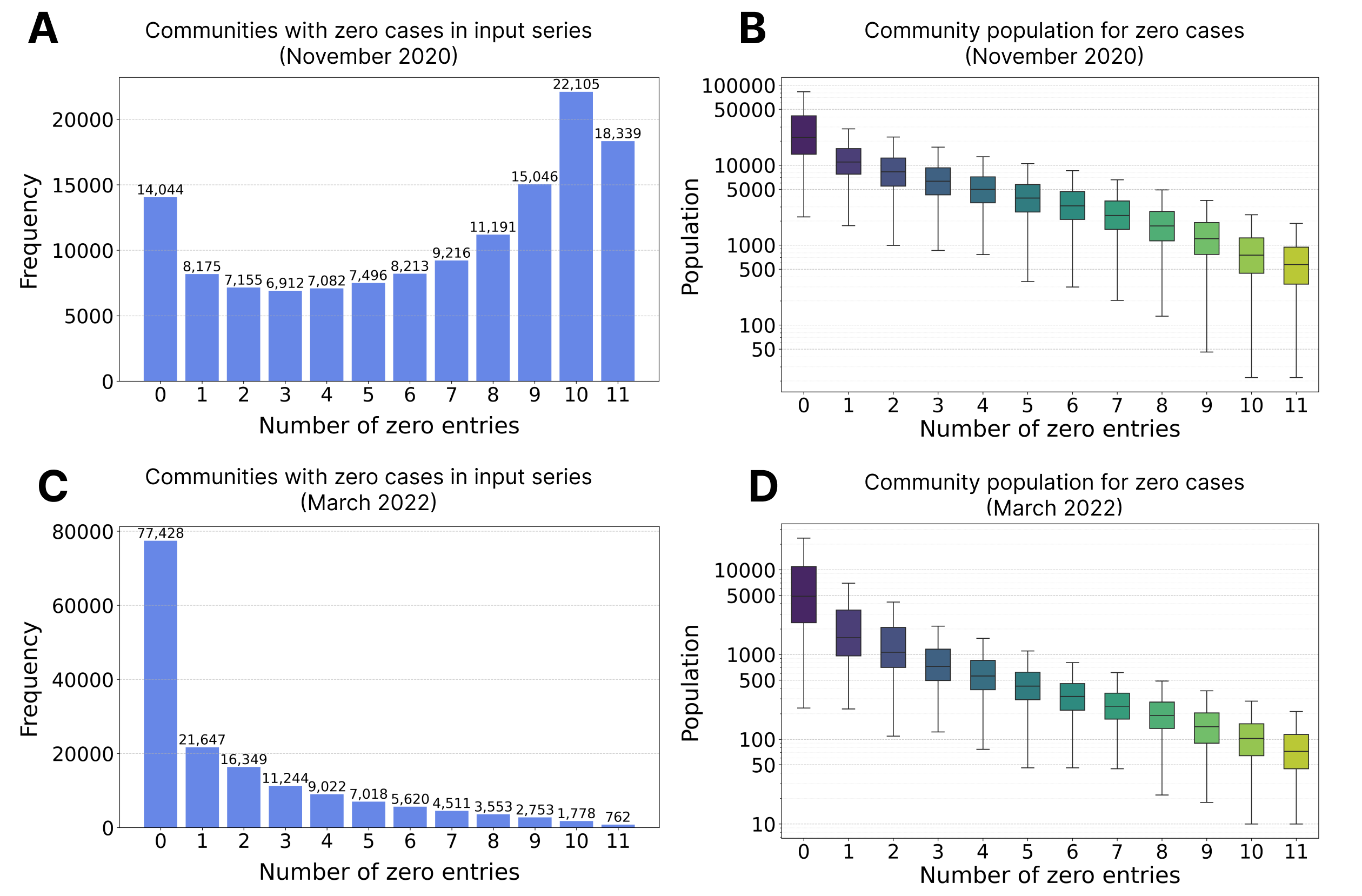}
    \caption{\textbf{Communities with zero cases on input horizon and prediction day.} Panel A shows the number of communities with 0 to 11 zero entries in the considered time series for November 2020 and Panel B shows the size of the community populations with the corresponding number of zeros. Panel C and D report the same structure for March 2022.}
    \label{fig:communities}
\end{figure}

In~\cref{fig:communities}, we provide the number of communities with zero confirmed cases over the input horizon $H=10$ days plus the one prediction day; cf.~Eqs.~\eqref{eq:input} and~\eqref{eq:output}. While we see that for March 2022 much less communities have a high number of zero entries, the approach showed similar problems for both periods. With the application of a 7-day centered moving average, we substantially reduced the number of zero entries but did not obtain a dataset for which we observed substantially different results. That means, we did not observe a prediction deterioration when the noise increased with $\varepsilon\rightarrow 0.3$. We suspect that this outcome is an artifact due to two properties. In the synthetically created dataset, we obtain either relatively many zero entries for a large number of community's time series or very small entries in the case of a moving average applied. In addition, the random sampling for the community dataset could have yielded time series which were already highly noisy such that the addition of DP noise did not qualitatively change the time series pattern to be learned. In order for FL and DP to be validated on community-level, we conclude that additional experiments will be necessary and that access to a true community-based dataset will be required.

\subsubsection*{Published county data}
\begin{figure}[h!]
    \begin{adjustwidth}{-2.25in}{0in}
        \centering
        \includegraphics[width=1.4\textwidth]{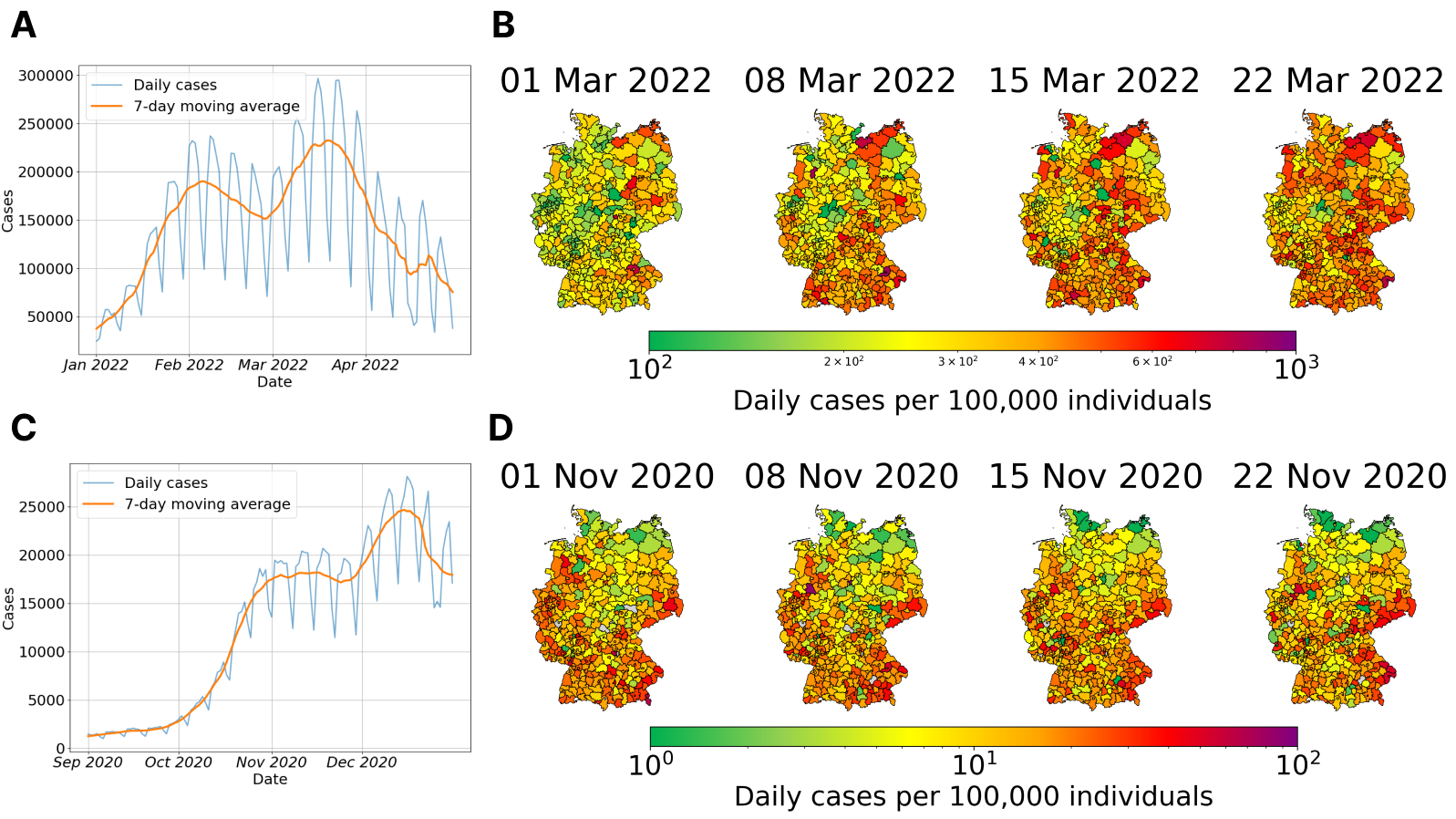}
        \caption{\textbf{Analysis of COVID-19 case numbers in Germany on county-level.} Panel A presents the time series of daily new infections throughout early 2022, showing both raw daily counts (blue) and the 7-day moving average (orange) that smooths the weekly reporting pattern. Panel B plots the geographical distribution of the COVID-19 incidence per 100,000 population at four weekly intervals in March 2022, illustrating the spatial heterogeneity of infection patterns and their temporal evolution. Panel C and D show the same type of data for late 2020.}
    	\label{fig:casedata}
    \end{adjustwidth}
\end{figure}

As county-level case data was the finest resolved data available, we validate our approach on this level. In~\cref{fig:casedata}, Panels A and C, we see the total infection dynamics in Germany with clearly visible waves of infection during early 2022 and late 2020. These Panels also demonstrate the typical weekly reporting pattern observed in the data, with pronounced drops in reported cases during weekends followed by increases during weekdays. To mitigate these reporting artifacts, we applied a centered 7-day moving average to the daily case counts, resulting in smoother time series as shown by the orange line.

In~\cref{fig:casedata}, Panels B and C, we present the geographical distribution of COVID-19 cases across the German counties. We see that the local entities have their own infection dynamics and that waves are partly desynchronized.

\subsection*{County-level prediction March 2022}
\begin{figure}[h!]
    \begin{adjustwidth}{-2.25in}{0in}
        \centering
        \includegraphics[width=\textwidth]{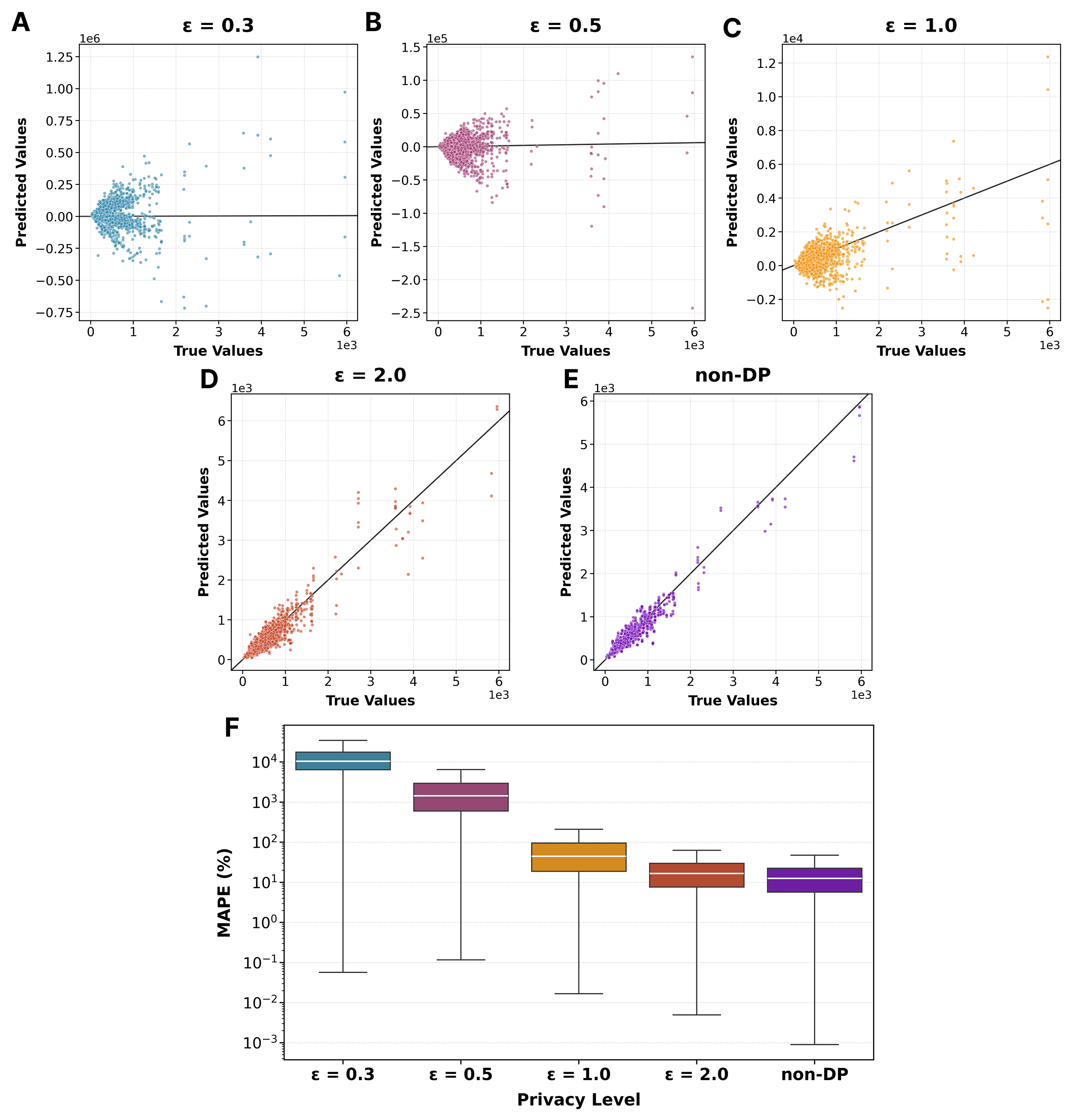}
        \caption{\textbf{Comparison of prediction performance across different privacy levels for March 2022 data}. Panels A-E show scatter plots of predicted against true case counts for various privacy budgets: (A) $\varepsilon = 0.3$, (B) $\varepsilon = 0.5$, (C) $\varepsilon = 1.0$, (D) $\varepsilon = 2.0$, and (E) $\varepsilon = \infty$ (non-DP). The black diagonal line represents perfect prediction. Panel F displays box plots of Mean Absolute Percentage Error (MAPE) for individual county predictions across all runs and privacy levels, illustrating the decreasing error and variance as privacy budget increases.}
    	\label{fig:2022_results}
    \end{adjustwidth}
\end{figure}

First, we evaluated our FL approach on county-level COVID-19 case data from March 2022, which represents a challenging period during the Omicron wave characterized by high peaks and dynamic infection trajectories. In~\cref{tab:2022_results} and~\cref{fig:2022_results}, we present the performance metrics across different privacy budget levels, ranging from strict privacy guarantees ($\varepsilon = 0.3$) to no differential privacy ($\varepsilon = \infty$, non-DP).

\begin{table}[h!]
\begin{adjustwidth}{-2.in}{0in}
\centering
\caption{\textbf{Performance metrics for COVID-19 case prediction on March 2022 data.} Results from 15 runs, 75 federated rounds, and 30 local epochs per round for different privacy budget levels. The metrics include Mean Squared Error (MSE), Mean Absolute Error (MAE), Mean Absolute Percentage Error (MAPE), and $R^2$ score. All values are reported as mean $\pm$ standard deviation across runs.}
\label{tab:2022_results}
\begin{tabular}{@{}l@{\hspace{3mm}}r@{\hspace{3mm}}r@{\hspace{3mm}}r@{\hspace{3mm}}r@{}}
\toprule
\textbf{$\varepsilon$} & \multicolumn{1}{c}{\textbf{MSE}} & \multicolumn{1}{c}{\textbf{MAE}} & \multicolumn{1}{c}{\textbf{MAPE (\%)}} & \multicolumn{1}{c}{\textbf{$R^2$}} \\
\midrule
0.3     & $1.26 \times 10^{10} \pm 1.65 \times 10^{10}$ & $6.80 \times 10^{4} \pm 4.69 \times 10^{4}$ & $1.33 \times 10^{4} \pm 9.02 \times 10^{3}$ & $-4.82 \times 10^{4} \pm 6.28 \times 10^{4}$ \\
0.5     & $2.99 \times 10^{8} \pm 3.98 \times 10^{8}$   & $9.97 \times 10^{3} \pm 8.17 \times 10^{3}$ & $1.96 \times 10^{3} \pm 1.60 \times 10^{3}$ & $-1.14 \times 10^{3} \pm 1.52 \times 10^{3}$ \\
1.0     & $4.00 \times 10^{5} \pm 5.71 \times 10^{5}$   & $349.55 \pm 299.23$     & $68.80 \pm 58.75$      & $-0.53 \pm 2.18$ \\
2.0     & $3.13 \times 10^{4} \pm 2.37 \times 10^{4}$   & $105.29 \pm 38.35$      & $20.75 \pm 6.30$       & $0.88 \pm 0.09$ \\
non-DP  & $1.91 \times 10^{4} \pm 1.54 \times 10^{3}$   & $81.42 \pm 3.16$        & $16.36 \pm 0.51$       & $0.93 \pm 0.01$ \\
\bottomrule
\end{tabular}
\end{adjustwidth}
\end{table}

Our results demonstrate a clear privacy-utility tradeoff that is characteristic of DP implementations. As shown in~\cref{fig:2022_results}, the quality of predictions declines substantially as we strengthen the privacy guarantees, i.e., by decreasing the privacy budget $\varepsilon$.  In particular, for $\varepsilon = 0.3$ and $\varepsilon = 0.5$ the models are effectively unusable as the predictions showing no correlation with the true case counts (see Panel A and B in~\cref{fig:2022_results}).
At the strictest privacy level ($\varepsilon = 0.3$), the model produced predictions that were almost completely unrelated to the actual values, with an extremely high MSE of $1.26 \times 10^{10}$ and a strongly negative $R^2$ value of approximately $-48208$. The mean absolute percentage error (MAPE) was over 13,000\%, which makes the complete failure of the predictive ability even more clear. Also, the standard deviation of these metrics across the runs is of the same order of magnitude or larger than the mean values themselves, indicating extreme instability in the prediction behavior.
At $\varepsilon = 0.5$, while still poor, the model begins to show marginal improvement, with MSE reduced by about two orders of magnitude to $2.99 \times 10^8$. However, the predictions remain basically unusable with MAPE near 2,000\% and $R^2$ still deeply negative at approximately $-1140$. \cref{fig:2022_results}(B) shows that while there is slightly more structure to the predictions, they still fail to track the true values in any meaningful way.
A substantial qualitative improvement emerges at $\varepsilon = 1.0$, where the MSE drops to around 400,000, and MAPE decreases to approximately 69\%. While the $R^2$ value remains slightly negative at $-0.53$, indicating the model still performs worse than a simple mean predictor, the predictions begin to show some correlation with the true values, as proven in~\cref{fig:2022_results}(C).
At $\varepsilon = 2.0$, we observe that the model achieves positive predictive value, with an $R^2$ of 0.88. Despite the privacy, the overall performance is comparable to the non-DP case. The MAPE decreases to around 21\%, and the predictions visibly track the true values as shown in~\cref{fig:2022_results}(D).
The non-private (non-DP) model shows our performance ceiling with the highest $R^2$ value of 0.93 and the lowest MAPE at 16.36\%. \cref{fig:2022_results}(E) demonstrates the most accurate tracking of true values.

Beyond the mean performance metrics, the standard deviations across runs reveal important insights about the stability of differentially private models. As $\varepsilon$ decreases, the standard deviations increase dramatically, reflecting the noise introduced by the privacy mechanism. This is especially evident in~\cref{fig:2022_results}(F), which shows box plots of the MAPE calculated for each individual county predictions. For these box plots, the individual MAPE values for each county from all runs were collected and then visualized separately for each privacy level. Each box plot thus represents the distribution of prediction errors across all counties and runs for a given privacy level.

The substantial outliers for $\varepsilon = 0.3$ and $\varepsilon = 0.5$ illustrate the extreme variability in model performance at these privacy levels, whereas the non-DP and $\varepsilon = 2.0$ cases show much more consistent performance.

\begin{figure}[h!]
    \begin{adjustwidth}{-2.25in}{0in}
        \centering
        \includegraphics[width=18cm]{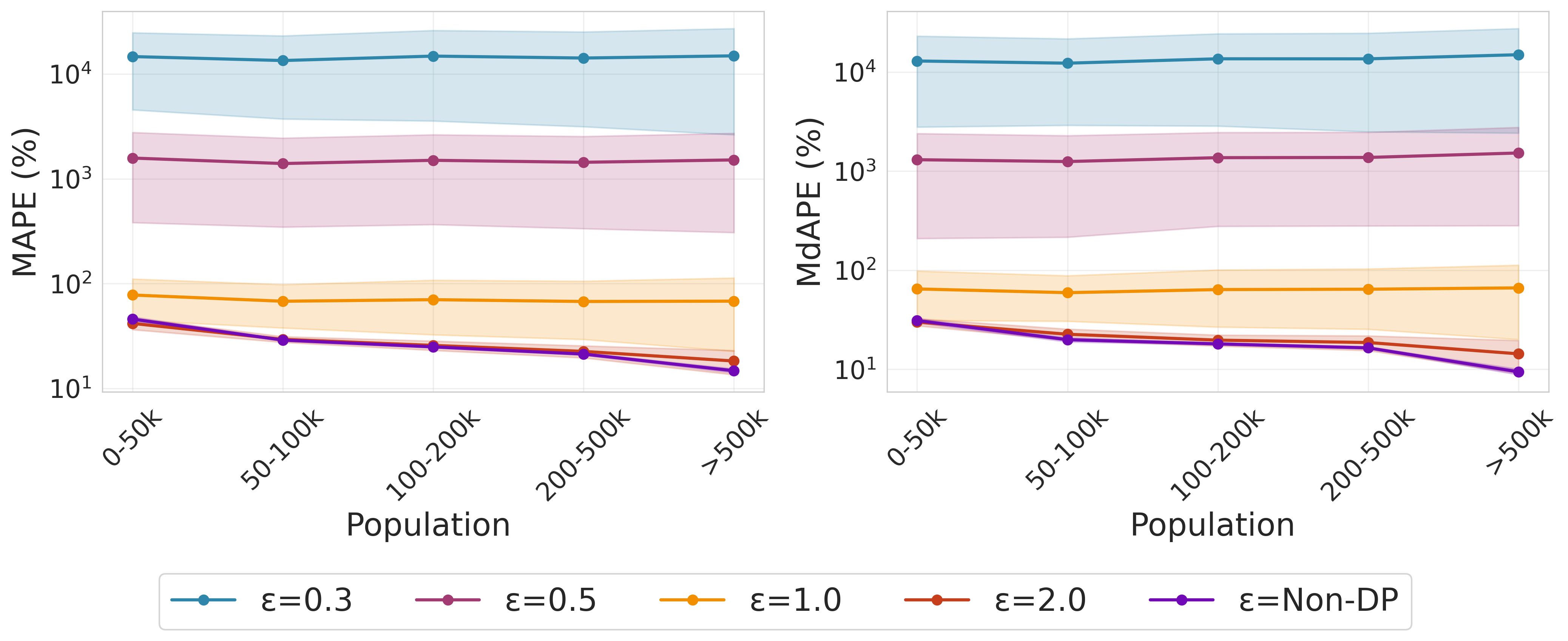}
        \caption{\textbf{Prediction perfomance stratified by county population size.} The panels show our metrics for different privacy levels ($\varepsilon \in \{0.3, 0.5, 1.0, 2.0\}$ and non-DP) across five population clusters (0--50k, 50--100k, 100--200k, 200--500k, $>$500k inhabitants) calculated from 15 runs, 75 federated rounds and 30 local epochs per round. The metrics include Mean Squared Error (MSE), Mean Absolute Error (MAE), Mean Absolute Percentage Error (MAPE), and $R^2$ score. The shaded areas represent the standard deviation across all 15 runs.}
        \label{fig:pop_eps}
        \end{adjustwidth}
    \end{figure}

To additionally analyze the impact of population heterogeneity, we stratified the test samples into five population clusters (0--50k, 50--100k, 100--200k, 200--500k, $>$500k). \Cref{fig:pop_eps} shows the relative performance metrics (MAPE and MdAPE) for each cluster. While absolute errors naturally scale with population size, these relative metrics mainly remain stable across all clusters. This suggests that smaller counties are not substantially disadvantaged by the global model. Instead, they likely benefit from the shared learning of temporal patterns driven by data-rich counties, which stabilizes predictions on their potential sparser local data.

These results highlight that for our dataset, there appears to be a minimum viable privacy budget (approximately $\varepsilon = 1.0$) below which predictions become too unreliable for practical use. However, the model at $\varepsilon = 2.0$ demonstrates that it is possible to nearly match the performance of a non-DP model while still providing strong privacy guarantees. This suggests that when designing privacy-preserving prediction systems for infectious disease dynamics, the balance of privacy requirements against the need for reliable predictions must be carefully considered.

\subsection*{County-level prediction November 2020}
In addition, we evaluate our model's performance and the DP tradeoff for an earlier stage of the pandemic as of November 2020. The November 2020 data evaluation reveals patterns consistent with those observed in the 2022 dataset, but with several notable differences that highlight how pandemic phase and disease dynamics influence the effectiveness of privacy-preserving predictive models. The metrics for the 2020 data are summarized in~\cref{tab:2020_results} and show a similar trend as the 2022 data in~\cref{tab:2022_results}.

\begin{figure}[h!]
    \begin{adjustwidth}{-2.25in}{0in}
        \centering
        \includegraphics[width=1.0\textwidth]{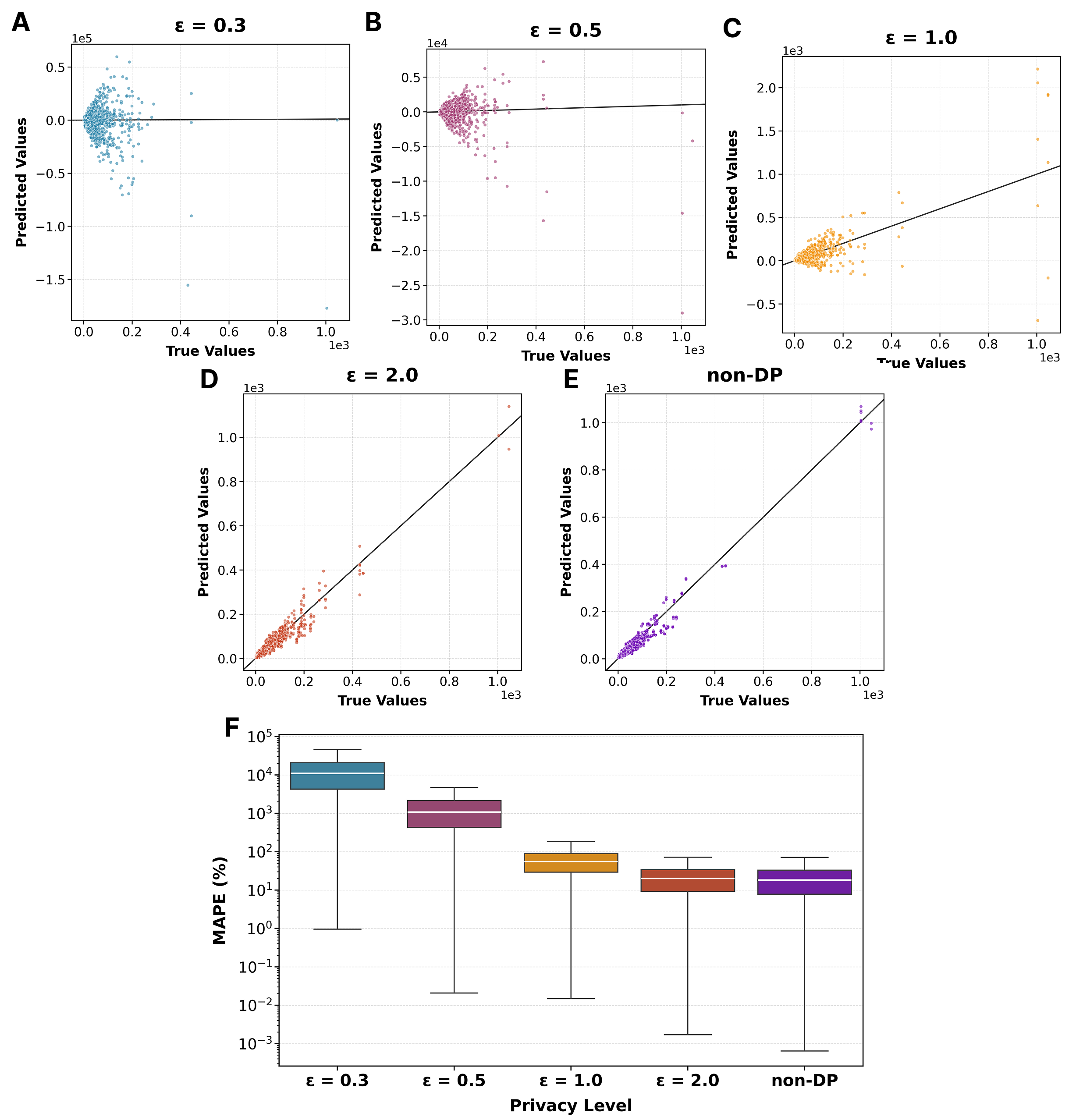}
        \caption{\textbf{Comparison of prediction performance across different privacy levels for November 2020 data}. Similar to Figure~\ref{fig:2022_results}, Panels A-E show scatter plots of predicted versus true case counts for various privacy budgets, while Panel F displays box plots of MAPE for individual county predictions across all runs.}
    	\label{fig:2020_results}
    \end{adjustwidth}
\end{figure}

Similar to the 2022 results, we observe a clear degradation in model performance as the privacy budget decreases. At $\varepsilon = 0.3$, the model again produces essentially unusable predictions with extremely high MSE ($2.03 \times 10^8$) and deeply negative $R^2$ values (approximately $-45940$). The MAPE values exceed 13,700\%, indicating complete prediction failure. As shown in~\cref{fig:2020_results}(A), predictions at this privacy level appear randomly distributed with no correlation to actual case counts.
At $\varepsilon = 0.5$, while still performing poorly, the model shows considerably better metrics than the equivalent privacy level in the 2022 data. The MSE drops to approximately 1.8 million, which represents a reduction of about two orders of magnitude compared to the 2022 results at the same privacy level ($2.99 \times 10^8$). The MAPE also shows relative improvement at 1,356\% versus 1,964\% in the 2022 data.
At $\varepsilon = 1.0$, the model achieves near-zero $R^2$ ($-0.01$), indicating performance approximately equivalent to a mean-value predictor. This is substantially better than the corresponding 2022 model at the same privacy level ($R^2 = -0.53$). The MSE improves to 4,480, compared to nearly 400,000 in the 2022 data. \cref{fig:2020_results}(C) shows predictions that begin to correlate with true values, but still with considerable dispersion.
At $\varepsilon = 2.0$, the model achieves good performance with $R^2 = 0.94$, nearly matching the non-DP model. The MAE of 9.37 is only marginally higher than the non-DP value of 8.52, suggesting that at this privacy level, the added noise has minimal impact on practical prediction quality. \cref{fig:2020_results}(D) shows predictions that closely track the actual values across the entire range of case counts.
The non-DP model establishes a slightly higher performance ceiling than in the 2022 data, with $R^2 = 0.95$ compared to 0.93. Also, the absolute errors are substantially lower in the 2020 data, with MAE of 8.52 against 81.42 in 2022, reflecting the lower absolute case numbers during this earlier pandemic phase. However, the MAPE is actually higher at 24.97\% versus 16.36\% in 2022, suggesting that despite lower absolute errors, the relative prediction accuracy may be somewhat worse.

The variance in model performance across multiple runs, as reflected in the standard deviations, follows similar patterns to the 2022 data, with extreme variability at low privacy budgets that stabilizes as $\varepsilon$ increases. \cref{fig:2020_results}(F) illustrates this through the MAPE box plots, which shows the distribution of individual percentage errors (MAPE) for each county across all runs. These box plots highlight the substantial outliers and wide distributions at $\varepsilon = 0.3$ and $\varepsilon = 0.5$ that contract substantially at higher privacy budgets.

In summary, the results for 2020 confirm the results from the 2022 data. Again, we were able to achieve good prediction performance at $\varepsilon = 2.0$, with the model closely matching the non-DP case but still preserving privacy guarantees.

\begin{table}[h!]
\begin{adjustwidth}{-2.in}{0in}
\centering
\caption{\textbf{Performance metrics for COVID-19 case prediction on November 2020 data.} Results from 15 runs, 75 federated rounds, and 30 local epochs per round.}
\label{tab:2020_results}
\begin{tabular}{@{}l@{\hspace{3mm}}r@{\hspace{3mm}}r@{\hspace{3mm}}r@{\hspace{3mm}}r@{}}
\toprule
\textbf{$\varepsilon$} & \multicolumn{1}{c}{\textbf{MSE}} & \multicolumn{1}{c}{\textbf{MAE}} & \multicolumn{1}{c}{\textbf{MAPE (\%)}} & \multicolumn{1}{c}{\textbf{$R^2$}} \\
\midrule
0.3 & $2.03 \times 10^8 \pm 2.91 \times 10^8$ & $5.77 \times 10^{3} \pm 5.00 \times 10^{3}$ & $1.37 \times 10^{4} \pm 1.14 \times 10^{4}$ & $-4.59 \times 10^{4} \pm 6.58 \times 10^{4}$ \\
0.5 & $1.80 \times 10^{6} \pm 2.53 \times 10^{6}$ & $566.38 \pm 454.17$ & $1.36 \times 10^{3} \pm 1.07 \times 10^{3}$ & $-406.51 \pm 574.01$ \\
1.0 & $4.48 \times 10^{3} \pm 4.36 \times 10^{3}$ & $30.84 \pm 16.44$ & $72.28 \pm 34.87$ & $-0.01 \pm 0.99$ \\
2.0 & $282.48 \pm 108.03$ & $9.37 \pm 1.37$ & $25.95 \pm 3.40$ & $0.94 \pm 0.02$ \\
non-DP & $213.14 \pm 4.55$ & $8.52 \pm 0.11$ & $24.97 \pm 0.55$ & $0.95 \pm 0.01$ \\
\bottomrule
\end{tabular}
\end{adjustwidth}
\end{table}

\subsection*{Privacy-utility tradeoff across epsilon values}
\begin{figure}[h!]
    \begin{adjustwidth}{-2.25in}{0in}
    \centering
    \includegraphics[width=18cm]{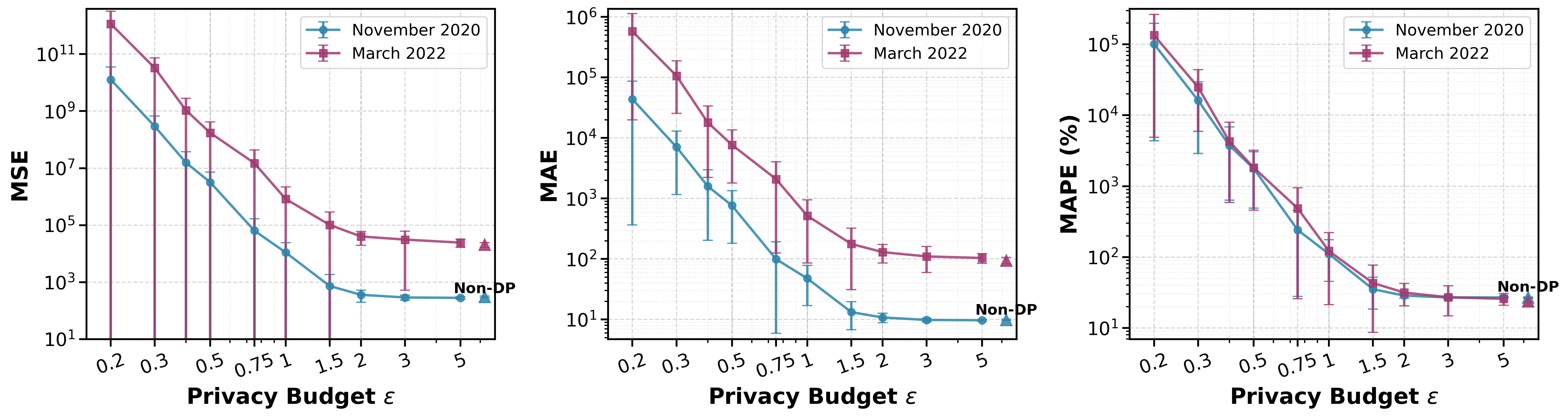}
    \caption{\textbf{Privacy-utility tradeoff across privacy budgets.} Privacy-utility curves for the November~2020 and March~2022 county-level datasets. Each panel shows mean $\pm$ standard deviation over $15$ runs of (A) MSE, (B) MAE, and (C) MAPE as a function of the privacy budget~$\varepsilon$. Both axes are on a logarithmic scale, and the non-DP baseline is indicated at the right-hand side of each curve.}
    \label{fig:epsilon_grid}
    \end{adjustwidth}
\end{figure}

 To provide a more explicit view of the privacy-utility tradeoff, we conduct additional experiments on county-level data for both epidemic periods with a finer grid of privacy budgets,
\begin{align*}
    \varepsilon\in\{0.2,0.3,0.4,0.5,0.75,1.0,1.5,2.0,3.0,5.0,\infty\}    
\end{align*}
We excluded $\varepsilon = 0.1$, because the RDP accountant produces prohibitively large noise multipliers at this extreme privacy level, which leads to completely unstable training.
For each value of $\varepsilon$ we repeated the DP-FL training $15$ times and recorded mean $\pm$ standard deviation of MSE, MAE, MAPE, and $R^2$.
The resulting privacy-utility curves in~\cref{fig:epsilon_grid} plot these metrics as a function of the privacy budget on logarithmic axes for November~2020 and March~2022.
For very strict privacy ($\varepsilon \le 0.5$), the forecasts are essentially unusable, with MAPE values exceeding $1{,}000~\%$ and very large variability across runs.
Between $\varepsilon = 0.5$ and $\varepsilon = 1.0$ the curves exhibit a sharp drop in all three error measures, reflecting the strongly nonlinear dependence of the RDP noise multiplier on $\varepsilon$.
For moderately strong privacy ($\varepsilon \geq 2$), the DP-FL models in both periods closely approach the non-DP baseline and additional increases of $\varepsilon$ yield only marginal utility gains.
The detailed summary statistics for all privacy budgets and both epidemic phases are reported in~\nameref{S1_tab}.

 \subsection*{Convergence across federated rounds}
        To analyze how predictive performance evolves during federated training under different privacy guarantees, we track the test error as a function of the number of federated rounds. For each epidemic period (November~2020 and March~2022) and for privacy budgets $\varepsilon \in \{0.5,1.0,2.0\}$, as well as the non-DP baseline, we train the model for $75$ federated rounds with the same settings as in the main experiments (30 local epochs per round, clipping norm $S = 0.5$, and $\delta = 10^{-5}$). Every 5 rounds, we evaluate the current global model on the same test set and record the MAPE. We repeat this for 15 runs and report mean $\pm$ standard deviation across runs.
        
\begin{figure}[H]
          \centering
          \includegraphics[width=\textwidth]{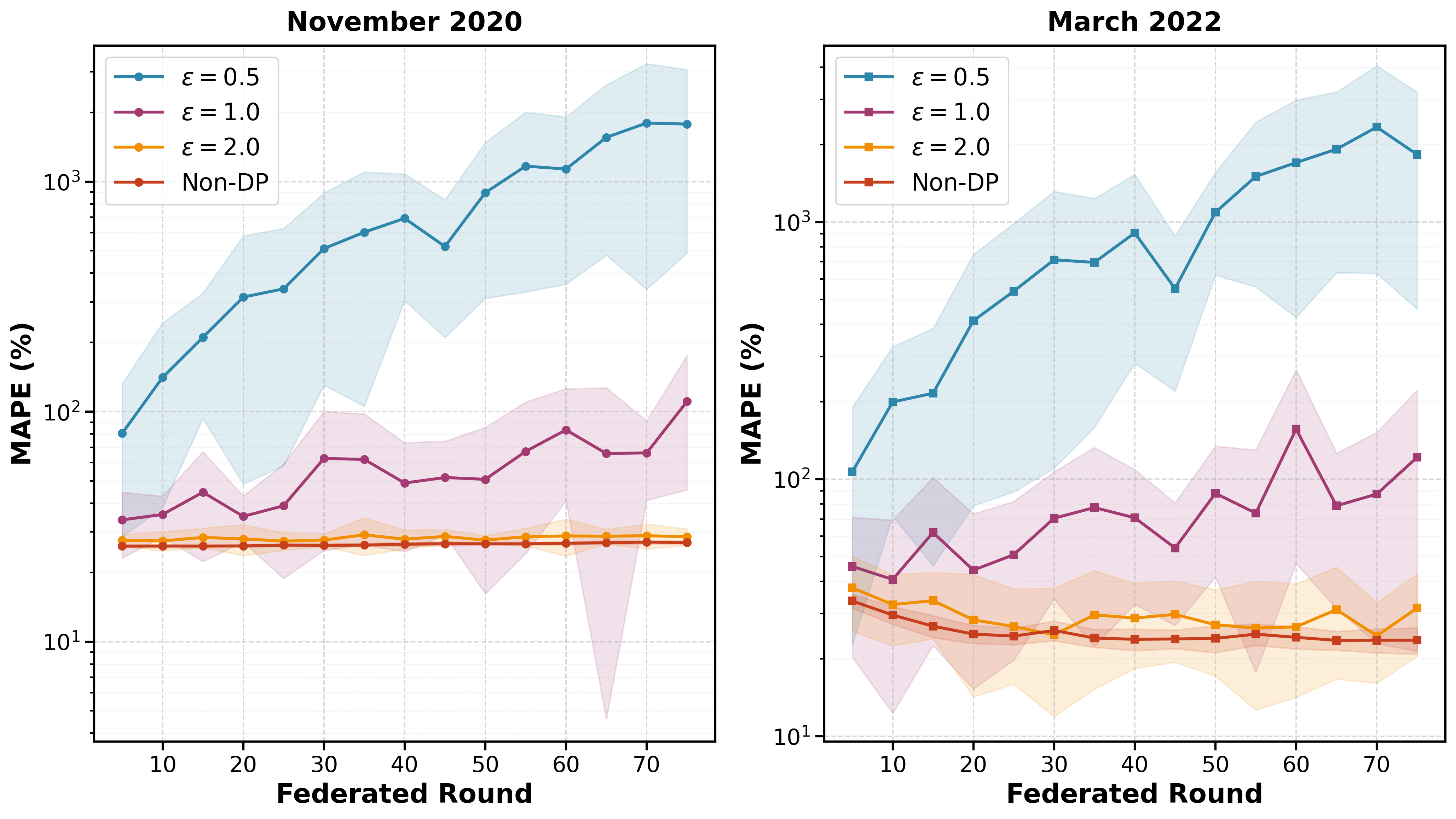}
          \caption{\textbf{Convergence of DP-FL across federated rounds for different privacy budgets.} Test MAPE (\%) as a function of the number of federated rounds for November~2020 (left) and March~2022 (right). Curves show mean $\pm$ standard deviation over 15 runs for privacy budgets $\varepsilon = 0.5$ (blue), $\varepsilon = 1.0$ (purple), $\varepsilon = 2.0$ (orange), and Non-DP (red).}
          \label{fig:convergence_rounds}
        \end{figure}

        \Cref{fig:convergence_rounds} shows the resulting curves. For very strict privacy ($\varepsilon = 0.5$), the test MAPE is already very high after a few rounds (around $80$-$100~\%$) and continues to increase over the course of training, with wide uncertainty bands that reflect large variability across runs. This indicates that the strong DP noise at this privacy level prevents stable learning and leads to increasingly noisy predictions. For intermediate privacy ($\varepsilon = 1.0$), the MAPE initially stabilizes (November~2020) or decreases slightly (March~2022) in the early rounds, but then stays at a much higher level than for $\varepsilon = 2.0$ (typically between $50$ and $150~\%$) and shows noticeable variation across runs. This suggests that additional rounds provide only small improvements and can even worsen performance when training is continued for too long.
        
        In contrast, for $\varepsilon = 2.0$ the curves quickly stabilize on a plateau with low MAPE values (about $25$-$30~\%$ in November~2020 and about $20$-$30~\%$ in March~2022). The non-DP baseline converges slightly faster and to slightly lower final MAPE values (around $25~\%$ in November~2020 and about $16$-$20~\%$ in March~2022), but the gap to $\varepsilon = 2.0$ remains small in both epidemic phases. The small increase of the $\varepsilon = 2.0$ curve at the last evaluation point in March~2022 lies well within the uncertainty variability band. Overall, these convergence patterns support the conclusion that privacy budgets around $\varepsilon \approx 2$ allow the model to reach useful forecasts, whereas very strict privacy ($\varepsilon = 0.5$) leads to unstable training dynamics and unusable predictions.
        
\section*{Discussion}
We developed a differential privacy-preserving federated learning approach for predicting COVID-19 case numbers on a spatially resolved level and validated the approach on county-level for two different time periods, November 2020 and March 2022. This was done to explore the balance between prediction accuracy and privacy-preservation in different stages of the pandemic. Our results demonstrate a clear relationship between privacy level (expressed through the privacy budget $\varepsilon$) and prediction quality.

One goal of our study was to explore the possibility of localized predictions at a finer spatial resolution than the county-level.  Although we developed a probabilistic approach to disaggregate county data to the community-level, the resulting data proved inadequately representative of actual local dynamics. The random assignment of cases to communities based solely on population data likely disrupted local spread patterns, resulting in time series dominated by unstructured noise rather than learnable transmission dynamics. In contrast, real-world community data, while noisy, retains the causal structure of infection chains which FL can leverage.
This experience underscores the challenges of working with sensitive health data at a highly resolved spatial level. While protecting privacy is a key concern, the availability of granular data is crucial for precise localized predictions. Due to the lack of case data resolved on community-level with often only several hundreds of individuals, we could not test our approach on very small population units. Future research will thus consider the creation of a more realistic synthetic dataset. This dataset should follow the structural patterns as observed during the COVID-19 pandemic on community-level. Furthermore, it should be created such that it can be published for scientific research. To do so, we will reach out to public health experts with access to more finely resolved data series. 

Our findings on county-level confirm the fundamental tradeoff between privacy and utility that is characteristic of differential privacy implementations. At very strict privacy guarantees ($\varepsilon = 0.3$ and $\varepsilon = 0.5$), predictions are practically unusable, as evidenced by extremely high MSE values and negative $R^2$ values. This observation aligns with the theoretical foundation of differential privacy, where a lower $\varepsilon$ value leads to stronger noise addition to model parameters.

For a privacy-preserving budget of $\varepsilon = 2.0$, we were able to achieve predictions that closely match the quality of non-DP models, with an $R^2$ value of 0.88 for the 2022 data and 0.94 for the 2020 data. The MAPE was around 21\% for the 2022 data and 25.95\% for the 2020 data, indicating that while there is some degradation in prediction quality compared to non-DP models, the results are still practically useful. This suggests that with a carefully chosen privacy budget, it is possible to achieve a good balance between privacy protection and prediction accuracy.

German counties vary substantially in population size (from under 50,000 to over 3.5 million inhabitants), which raises concerns about whether smaller counties might be disadvantaged in a federated learning setting. To address this, we conducted an analysis stratifying model performance by population clusters. The results show that while smaller counties exhibit higher absolute errors due to lower case counts, the privacy-utility tradeoff remains relatively uniform across all population sizes. Importantly, smaller counties benefit from the global model by learning temporal patterns that would be undetectable in their local data alone. The gradient clipping mechanism in our DP-FL algorithm ($S=0.5$) prevents larger counties from dominating the aggregation, ensuring that updates from all counties contribute equally.

We base our experiments on an established client-level DP-FL setup to isolate and interpret the impact of the privacy budget on forecast quality. Prior work has proposed a range of refinements, including local-DP federated learning with per-client privacy modules~\cite{10.1145/3378679.3394533}, algorithms that study different locations for adding noise and analyze their convergence and utility guarantees~\cite{9069945}, and healthcare-focused FL frameworks~\cite{kerkouche2021privacy,rahimian2022practical,ryu_appfl_2022,madduri24}. These methods target algorithmic improvements and can in principle be combined with our setting. We expect that these algorithms may further improve forecast accuracy, a systematic evaluation of such variants for infectious disease dynamics should be included in future work. Regarding the model choice, we utilized a standard MLP to demonstrate the impact of the privacy mechanism; however, the framework is compatible with any gradient-based model. We consider this work as a proof of concept with more work required on the way to this methodology being used as designated. 

A limitation of our study concerns the quality and resolution of the underlying data. Clearly, it would have been better if we could also have validated our approach on the much smaller communities. Besides that we attempted to mitigate weekend effects by applying a 7-day moving average, these partially remain and limit the models learning ability. Eventually, problem of underreporting prevails: the reported case numbers represent only a fraction of actual infections with differences in spatial levels. According to data from the Robert Koch Institute~\cite{Robert_Koch-Institut_SARS-CoV-2_Infektionen_in_2025}, test positivity rates in 2022 ranged between 30~\% and 50~\%, suggesting a substantial dark figure. However, we assume that the underreporting rate remains relatively constant within the short timeframes considered. Future research should incorporate auxiliary data sources, such as wastewater-based indicators or dark figure estimators, to better account for true infection numbers.

The tradeoff between the need for detailed data for precise predictions and privacy requirements underscores the relevance of our FL approach with DP, which can enable LHAs to use their granular data for collaborative learning without directly sharing it.

As an important outcome for public health, we conclude that with an appropriate privacy budget ($\varepsilon = 2.0$), DP-FL models can generate predictions that nearly match the quality of non-DP models. This opens up possibilities for privacy-compliant collaboration between LHAs without heavily affecting the quality of decision support. However, it should be noted that privacy budgets might need dynamic adaptation to the epidemic situation. 

Overall, our work demonstrates that privacy-preserving federated learning approaches are suitable in principle for predicting infectious disease dynamics if the privacy budget is carefully chosen.

\section*{Supporting information}

\paragraph*{S1 Tab.}  Performance metrics across a fine grid of privacy budgets.
\label{S1_tab}

\paragraph*{S1 Text.} Formal derivation of the privacy guarantees using the Gaussian Mechanism and R\'enyi Differential Privacy (RDP). The appendix includes definitions, the privacy accounting methodology, and the conversion theorems used to establish the final $(\varepsilon, \delta)$-differential privacy bounds.
\label{S1_text}





\section*{Acknowledgments}
This work was supported by the Initiative and Networking Fund of the Helmholtz Association (grant agreement number KA1-Co-08, Project LOKI-Pandemics). The funders had no role in study design, data collection and analysis, decision to publish, or preparation of the manuscript.

\section*{Competing interests}
The authors declare to not have any competing interests.

\section*{Data availability}
All data used in this study are publicly available from the Robert Koch Institute (RKI) and can be accessed via the \textit{memilio-epidata} Python package~\cite{Bicker_MEmilio_v2_0_0_-}. The code used to generate the results of this study is available at \url{https://github.com/SciCompMod/memilio-simulations}.

\section*{Author Contributions}
    \noindent\textbf{Conceptualization:} All authors\\
    \noindent\textbf{Data Curation:} Henrik Zunker, Raouf Kerkouche\\
    \noindent\textbf{Formal Analysis:} Henrik Zunker, Raouf Kerkouche, Martin Kühn\\
    \noindent\textbf{Funding Acquisition:} Mario Fritz, Martin Kühn\\
    \noindent\textbf{Investigation:} Henrik Zunker, Raouf Kerkouche, Martin Kühn\\
    \noindent\textbf{Methodology:} All authors\\
    \noindent\textbf{Project Administration:} Mario Fritz, Martin Kühn\\
    \noindent\textbf{Resources:} Mario Fritz, Martin Kühn\\
    \noindent\textbf{Software:} Henrik Zunker, Raouf Kerkouche \\
    \noindent\textbf{Supervision:}  Mario Fritz, Martin Kühn\\
    \noindent\textbf{Validation:} All authors \\
    \noindent\textbf{Visualization:} Henrik Zunker \\
    \noindent\textbf{Writing – Original Draft:} Henrik Zunker, Raouf Kerkouche\\
    \noindent\textbf{Writing – Review \& Editing:} All authors


%
%
%

{
\newpage
\section*{S1.Appendix: Performance metrics across privacy budgets}
\begin{table}[H]
    \centering
    \caption*{\textbf{S1 Tab. Performance metrics across privacy budgets.} Results from 15 runs and 75 federated rounds for different $\varepsilon$ values. All values are mean $\pm$ standard deviation.}
    \label{tab:epsilon_grid}
    \resizebox{\textwidth}{!}{%
    \begin{tabular}{llrrrr}
    \toprule
    Year & $\varepsilon$ & \multicolumn{1}{c}{MSE} & \multicolumn{1}{c}{MAE} & \multicolumn{1}{c}{MAPE (\%)} & \multicolumn{1}{c}{R2} \\
    \midrule
    Nov 2020 & 0.20 & $1.28\mathrm{e}{+10} \pm 2.22\mathrm{e}{+10}$ & $4.38\mathrm{e}{+04} \pm 4.34\mathrm{e}{+04}$ & $1.01\mathrm{e}{+05} \pm 9.67\mathrm{e}{+04}$ & $-2.74\mathrm{e}{+06} \pm 4.76\mathrm{e}{+06}$ \\
    Nov 2020 & 0.30 & $2.91\mathrm{e}{+08} \pm 3.76\mathrm{e}{+08}$ & $7.08\mathrm{e}{+03} \pm 5.92\mathrm{e}{+03}$ & $1.62\mathrm{e}{+04} \pm 1.33\mathrm{e}{+04}$ & $-6.23\mathrm{e}{+04} \pm 8.05\mathrm{e}{+04}$ \\
    Nov 2020 & 0.40 & $1.55\mathrm{e}{+07} \pm 2.17\mathrm{e}{+07}$ & $1.60\mathrm{e}{+03} \pm 1.40\mathrm{e}{+03}$ & $3.75\mathrm{e}{+03} \pm 3.12\mathrm{e}{+03}$ & $-3.32\mathrm{e}{+03} \pm 4.64\mathrm{e}{+03}$ \\
    Nov 2020 & 0.50 & $3.17\mathrm{e}{+06} \pm 4.07\mathrm{e}{+06}$ & $7.63\mathrm{e}{+02} \pm 5.81\mathrm{e}{+02}$ & $1.78\mathrm{e}{+03} \pm 1.29\mathrm{e}{+03}$ & $-6.79\mathrm{e}{+02} \pm 8.71\mathrm{e}{+02}$ \\
    Nov 2020 & 0.75 & $6.41\mathrm{e}{+04} \pm 1.02\mathrm{e}{+05}$ & $9.96\mathrm{e}{+01} \pm 9.37\mathrm{e}{+01}$ & $2.41\mathrm{e}{+02} \pm 2.14\mathrm{e}{+02}$ & $-1.27\mathrm{e}{+01} \pm 2.19\mathrm{e}{+01}$ \\
    Nov 2020 & 1.00 & $1.10\mathrm{e}{+04} \pm 1.31\mathrm{e}{+04}$ & $4.77\mathrm{e}{+01} \pm 3.07\mathrm{e}{+01}$ & $1.11\mathrm{e}{+02} \pm 6.50\mathrm{e}{+01}$ & $-1.36\mathrm{e}{+00} \pm 2.80\mathrm{e}{+00}$ \\
    Nov 2020 & 1.50 & $7.32\mathrm{e}{+02} \pm 1.13\mathrm{e}{+03}$ & $1.32\mathrm{e}{+01} \pm 6.49\mathrm{e}{+00}$ & $3.52\mathrm{e}{+01} \pm 1.68\mathrm{e}{+01}$ & $8.43\mathrm{e}{-01} \pm 2.42\mathrm{e}{-01}$ \\
    Nov 2020 & 2.00 & $3.59\mathrm{e}{+02} \pm 1.65\mathrm{e}{+02}$ & $1.08\mathrm{e}{+01} \pm 1.88\mathrm{e}{+00}$ & $2.86\mathrm{e}{+01} \pm 2.33\mathrm{e}{+00}$ & $9.23\mathrm{e}{-01} \pm 3.54\mathrm{e}{-02}$ \\
    Nov 2020 & 3.00 & $2.90\mathrm{e}{+02} \pm 5.80\mathrm{e}{+01}$ & $9.83\mathrm{e}{+00} \pm 6.82\mathrm{e}{-01}$ & $2.69\mathrm{e}{+01} \pm 1.67\mathrm{e}{+00}$ & $9.38\mathrm{e}{-01} \pm 1.24\mathrm{e}{-02}$ \\
    Nov 2020 & 5.00 & $2.81\mathrm{e}{+02} \pm 3.12\mathrm{e}{+01}$ & $9.70\mathrm{e}{+00} \pm 3.39\mathrm{e}{-01}$ & $2.69\mathrm{e}{+01} \pm 5.65\mathrm{e}{-01}$ & $9.40\mathrm{e}{-01} \pm 6.70\mathrm{e}{-03}$ \\
    Nov 2020 & Non-DP & $3.01\mathrm{e}{+02} \pm 2.40\mathrm{e}{+01}$ & $9.85\mathrm{e}{+00} \pm 2.71\mathrm{e}{-01}$ & $2.70\mathrm{e}{+01} \pm 3.65\mathrm{e}{-01}$ & $9.36\mathrm{e}{-01} \pm 5.10\mathrm{e}{-03}$ \\
    \midrule
    Mar 2022 & 0.20 & $1.16\mathrm{e}{+12} \pm 2.07\mathrm{e}{+12}$ & $5.77\mathrm{e}{+05} \pm 5.58\mathrm{e}{+05}$ & $1.34\mathrm{e}{+05} \pm 1.30\mathrm{e}{+05}$ & $-6.80\mathrm{e}{+06} \pm 1.22\mathrm{e}{+07}$ \\
    Mar 2022 & 0.30 & $3.25\mathrm{e}{+10} \pm 4.14\mathrm{e}{+10}$ & $1.07\mathrm{e}{+05} \pm 8.12\mathrm{e}{+04}$ & $2.49\mathrm{e}{+04} \pm 1.90\mathrm{e}{+04}$ & $-1.91\mathrm{e}{+05} \pm 2.44\mathrm{e}{+05}$ \\
    Mar 2022 & 0.40 & $1.07\mathrm{e}{+09} \pm 1.76\mathrm{e}{+09}$ & $1.81\mathrm{e}{+04} \pm 1.59\mathrm{e}{+04}$ & $4.28\mathrm{e}{+03} \pm 3.69\mathrm{e}{+03}$ & $-6.27\mathrm{e}{+03} \pm 1.04\mathrm{e}{+04}$ \\
    Mar 2022 & 0.50 & $1.72\mathrm{e}{+08} \pm 2.42\mathrm{e}{+08}$ & $7.69\mathrm{e}{+03} \pm 5.89\mathrm{e}{+03}$ & $1.83\mathrm{e}{+03} \pm 1.37\mathrm{e}{+03}$ & $-1.01\mathrm{e}{+03} \pm 1.42\mathrm{e}{+03}$ \\
    Mar 2022 & 0.75 & $1.50\mathrm{e}{+07} \pm 2.88\mathrm{e}{+07}$ & $2.09\mathrm{e}{+03} \pm 1.97\mathrm{e}{+03}$ & $4.87\mathrm{e}{+02} \pm 4.61\mathrm{e}{+02}$ & $-8.71\mathrm{e}{+01} \pm 1.69\mathrm{e}{+02}$ \\
    Mar 2022 & 1.00 & $8.31\mathrm{e}{+05} \pm 1.34\mathrm{e}{+06}$ & $5.17\mathrm{e}{+02} \pm 4.31\mathrm{e}{+02}$ & $1.22\mathrm{e}{+02} \pm 1.00\mathrm{e}{+02}$ & $-3.88\mathrm{e}{+00} \pm 7.91\mathrm{e}{+00}$ \\
    Mar 2022 & 1.50 & $1.02\mathrm{e}{+05} \pm 1.86\mathrm{e}{+05}$ & $1.77\mathrm{e}{+02} \pm 1.46\mathrm{e}{+02}$ & $4.30\mathrm{e}{+01} \pm 3.43\mathrm{e}{+01}$ & $4.01\mathrm{e}{-01} \pm 1.10\mathrm{e}{+00}$ \\
    Mar 2022 & 2.00 & $4.00\mathrm{e}{+04} \pm 2.06\mathrm{e}{+04}$ & $1.30\mathrm{e}{+02} \pm 4.40\mathrm{e}{+01}$ & $3.16\mathrm{e}{+01} \pm 1.11\mathrm{e}{+01}$ & $7.65\mathrm{e}{-01} \pm 1.21\mathrm{e}{-01}$ \\
    Mar 2022 & 3.00 & $3.10\mathrm{e}{+04} \pm 3.05\mathrm{e}{+04}$ & $1.10\mathrm{e}{+02} \pm 5.05\mathrm{e}{+01}$ & $2.71\mathrm{e}{+01} \pm 1.23\mathrm{e}{+01}$ & $8.18\mathrm{e}{-01} \pm 1.80\mathrm{e}{-01}$ \\
    Mar 2022 & 5.00 & $2.44\mathrm{e}{+04} \pm 7.76\mathrm{e}{+03}$ & $1.04\mathrm{e}{+02} \pm 1.95\mathrm{e}{+01}$ & $2.58\mathrm{e}{+01} \pm 4.85\mathrm{e}{+00}$ & $8.57\mathrm{e}{-01} \pm 4.56\mathrm{e}{-02}$ \\
    Mar 2022 & Non-DP & $2.04\mathrm{e}{+04} \pm 3.74\mathrm{e}{+03}$ & $9.42\mathrm{e}{+01} \pm 1.14\mathrm{e}{+01}$ & $2.36\mathrm{e}{+01} \pm 2.78\mathrm{e}{+00}$ & $8.80\mathrm{e}{-01} \pm 2.20\mathrm{e}{-02}$ \\
    \bottomrule
    \end{tabular}}
\end{table}

\section*{S2.Appendix: Privacy analysis}
 \section*{Privacy Analysis}
    \label{sec:priv_analysis}

    The privacy analysis of our differentially private baseline is discussed here.
    As in~\cite{abadi2016deep,mironov2019r}, we consider the Sampled Gaussian Mechanism (SGM) —a composition of subsampling and the additive Gaussian noise (defined in Definition~\ref{def:sgm})— for privacy amplification. Moreover, we first compute the SGM’s Renyi Differential Privacy as in~\cite{mironov2019r} and then we use the conversion Theorem~\ref{th:conv} from~\cite{balle2020hypothesis} to switch back to Differential Privacy.

    \begin{definition}[R\'enyi divergence]    
    Let $P$ and $Q$ be two distributions on $\mathcal{X}$ defined over the same probability space, and let $p$ and $q$ be their respective densities. The R\'enyi divergence of a finite order $\alpha \neq 1$ between $P$ and $Q$ is defined as follows:
    
    \begin{equation}
        D_\alpha\left( P \parallel Q \right) \overset{\Delta}{=} \frac{1}{\alpha -1} \ln \int_{\mathcal{X}} q(x) \left( \frac{p(x)}{q(x)} \right)^\alpha \mathop{dx}. \nonumber
    \end{equation}
    
    R\'enyi divergence at orders $\alpha=1,\infty$ is defined by continuity.
    
    \end{definition}
    
    \vspace{10pt}
    
    \begin{definition}[R\'enyi differential privacy (RDP)]
    
    A randomized mechanism $\gM: \gE \rightarrow \gR$ satisfies $(\alpha, \rho)$-R\'enyi differential privacy (RDP) if for any two adjacent inputs $E$, $E' \in \gE$ it holds that
    
    \begin{equation}
      D_\alpha\left( \gM(D) \parallel \gM(D')  \right) \leq \rho   \nonumber
    \end{equation}
        
    \end{definition}
    
    In this work, we call two datasets $D, D'$ to be adjacent if $D'= D \cup \{ x \}$ (or vice versa).
    
    \vspace{10pt}
    \begin{definition}[Sampled Gaussian Mechanism (SGM)]
    \label{def:sgm}
        Let $f$ be an arbitrary function mapping subsets of $\mathcal{E}$ to $\mathbb{R}^d$. We define the Sampled Gaussian mechanism (SGM) parametrized with the sampling rate $0 < q \leq 1$ and the noise $\sigma>0$ as
        \begin{equation}
         \mathrm{SG}_{q,\sigma} \overset{\Delta}{=}  f \left(\{ x : x \in D \text{ is sampled with probability } q \} \right) + \gN(0,\sigma^2\mathbb{I}^d),\nonumber      
        \end{equation}
    where each element of $D$ is independently and randomly sampled  with probability $q$ without replacement. 
    
    As for the Gaussian Mechanism, the Sampled Gaussian Mechanism consists of adding identically and independently distributed Gaussian noise with zero mean and variance $\sigma^2$ to each coordinate value of the true output of $f$.
    In fact, the Sampled Gaussian Mechanism draws vector values from a
    multivariate spherical (or isotropic) Gaussian distribution which
    is described by random variable $\gN(0,\sigma^2\mathbb{I}^d)$, where $d$ is omitted if it is unambiguous in the given context.

    \end{definition}
    
    \subsubsection*{Analysis}
    The privacy guarantee of our approach is quantified using the revisited moment accountant~\cite{mironov2019r} that restates the moments accountant introduced in~\cite{abadi2016deep} using the notion of R\'enyi differential privacy (RDP) defined in~\cite{mironov2017renyi}. 
    
    Let $\mu_0$ denote the probability density function ($\mathrm{pdf}$) of $\mathcal{N}(0,\sigma^2)$ and let $\mu_1$ denote the $\mathrm{pdf}$ of $\mathcal{N}(1,\sigma^2)$. Let $\mu$ be the mixture of two Gaussians $\mu= (1-q)\mu_0 + a\mu_1$, where $q$ is the sampling probability of a single record in a single round.
    
    \begin{theorem}
    \label{th:mix_simple}
        Let $\mathrm{SG}_{q,\sigma}$ be the Sampled Gaussian mechanism for some function $f$ and under the assumption $\Delta_2 f \leq 1$ for any adjacent $D, D' \in \mathcal{E}$. Then $\mathrm{SG}_{q,\sigma}$ satisfies $(\alpha,\rho)$-RDP if
        \begin{equation}
        \label{eq:logmax}
        \rho \leq \frac{1}{\alpha-1} \log \max (A_{\alpha},B_{\alpha})
    \end{equation}
    
    where $A_{\alpha} \overset{\Delta}{=} \mathbb{E}_{z\sim \mu_0} [\left( \mu(z)/\mu_0(z)\right)^\alpha] $
      and $ B_{\alpha} \overset{\Delta}{=} \mathbb{E}_{z\sim \mu} [\left( \mu_0(z)/\mu(z)\right)^\alpha]$

    \end{theorem}

    Theorem~\ref{th:mix_simple} states that applying $\mathrm{SGM}$ to a function of sensitivity (Definition~\ref{def:sensitivity}) at most 1 (which also holds for larger values without loss of generality) satisfies $(\alpha,\rho)$-RDP if  $\rho \leq \frac{1}{\alpha-1} \log(\max \{A_{\alpha},B_{\alpha}\} )$. Thus, analyzing RDP properties of SGM is equivalent to upper bounding $A_{\alpha}$ and $B_{\alpha}$.
    
    From Corollary 7. in~\cite{mironov2019r}, $A_{\alpha}\geq B_{\alpha}$ for any $\alpha \geq 1$. Therefore, we can reformulate~\cref{eq:logmax} as
    
    \begin{equation}
        \label{eq:xi}
        \rho \leq \xi_{\mathcal{N}}(\alpha|q) \coloneqq \frac{1}{\alpha-1} \log A_{\alpha}
    \end{equation}
    
    To compute $A_{\alpha}$, we use the numerically stable computation approach proposed in~\cite{mironov2019r} (Sec. 3.3) depending on whether $\alpha$ is expressed as an integer or a real value.
    
    \begin{theorem}[Composability~\cite{mironov2017renyi}]
    \label{th:comp}
    Suppose that a mechanism $\mathcal{M}$ consists of a sequence of adaptive mechanisms $\mathcal{M}_1, \dots, \mathcal{M}_k$ where $\mathcal{M}_i: \prod_{j=1}^{i-1} \mathcal{R}_j \times \mathcal{E} \rightarrow \mathcal{R}_i$. If all the mechanisms in the sequence are $(\alpha,\rho)$-RDP, then the composition of the sequence is $(\alpha,k\rho)$-RDP.
    \end{theorem}
    
    In particular, Theorem~\ref{th:comp} holds when the mechanisms themselves are chosen based on the (public) output of the previous mechanisms. By Theorem~\ref{th:comp}, it suffices to compute $\xi_{\mathcal{N}}(\alpha|q)$ at each step and sum them up to bound the overall RDP privacy budget of an iterative mechanism composed of DP mechanisms obtained over different steps.

    \begin{theorem}[Conversion from RDP to DP~\cite{balle2020hypothesis}]
    \label{th:conv}
    If a mechanism $\mathcal{M}$ is $(\alpha,\rho)$-RDP then it is 
    $((\rho + \log((\alpha-1)/\alpha) - (\log \delta + \log \alpha)/(\alpha -1),\delta)$-DP for any $0<\delta<1$.
    
    \end{theorem}
    
    \begin{theorem}[Privacy of our approach]
    For any $0<\delta<1$ and $\alpha \geq 1$, our approach is \\ $\left( \min_{\alpha} (\Tcl \cdot \xi(\alpha|q) + \log((\alpha-1)/\alpha) - (\log \delta + \log \alpha)/(\alpha -1)), \delta  \right)$-DP, where  $\xi_{\mathcal{N}}(\alpha|q)$ is defined in~\cref{eq:xi}.
    \end{theorem}
    
    The proof follows from~\cref{th:mix_simple,th:comp,th:conv} and the fact that a client (LHA) is sampled in every federated round with a probability of $q$.

}

\end{document}